\documentclass[sigconf]{acmart}

\usepackage{enumitem}
\usepackage{graphicx}
\usepackage{fancyhdr}
\usepackage{amsmath}
\usepackage{amsthm}
\usepackage{amsfonts}
\usepackage{amssymb}
\usepackage{bm}
\usepackage{url}
\usepackage{hyperref}
\usepackage{booktabs}
\usepackage{multirow}
\usepackage{multicol}
\usepackage{color}
\usepackage{algorithm}
\usepackage{algpseudocode}
\usepackage{float}
\usepackage{calc}  
\usepackage{mathtools}
\usepackage{caption}
\usepackage{subcaption}
\usepackage{booktabs}
\usepackage{rotating}

\AtBeginDocument{%
  \providecommand\BibTeX{{%
    \normalfont B\kern-0.5em{\scshape i\kern-0.25em b}\kern-0.8em\TeX}}}

\copyrightyear{2020}
\acmYear{2020}
\setcopyright{acmcopyright}\acmConference[KDD '20]{Proceedings of the 26th ACM
SIGKDD Conference on Knowledge Discovery and Data Mining}{August 23--27,
2020}{Virtual Event, CA, USA}
\acmBooktitle{Proceedings of the 26th ACM SIGKDD Conference on Knowledge
Discovery and Data Mining (KDD '20), August 23--27, 2020, Virtual Event, CA, USA}
\acmPrice{15.00}
\acmDOI{10.1145/3394486.3403247}
\acmISBN{978-1-4503-7998-4/20/08}

\acmSubmissionID{rfp2027}

\newcommand{\Black}[1]{\textcolor[rgb]{0.00,0.00,0.00}{#1}}
\newcommand{\reviseyq}[1]{\Black{#1}}
\newcommand{\revisexin}[1]{\Black{#1}}

\begin{document}

\title{Neural Subgraph Isomorphism Counting}

\author{Xin Liu}
\email{xliucr@cse.ust.hk}
\affiliation{%
  \institution{Department of CSE}
  \institution{HKUST}
}

\author{Haojie Pan}
\email{hpanad@cse.ust.hk}
\affiliation{%
  \institution{Department of CSE}
  \institution{HKUST}
}

\author{Mutian He}
\email{mhear@cse.ust.hk}
\affiliation{%
  \institution{Department of CSE}
  \institution{HKUST}
}

\author{Yangqiu Song}
\email{yqsong@cse.ust.hk}
\affiliation{%
  \institution{Department of CSE}
  \institution{HKUST}
}

\author{Xin Jiang}
\email{jiang.xin@huawei.com}
\affiliation{%
  \institution{Huawei Noah's Ark Lab}
  \institution{Huawei Technologies Co. Ltd}
}

\author{Lifeng Shang}
\email{shang.lifeng@huawei.com}
\affiliation{%
  \institution{Huawei Noah's Ark Lab}
  \institution{Huawei Technologies Co. Ltd}
}


\begin{abstract}
In this paper, we study a new graph learning problem: learning to count subgraph isomorphisms. 
Different from other traditional graph learning problems such as node classification and link prediction, subgraph isomorphism counting is NP-complete and requires more global inference to oversee the whole graph.
To make it scalable for large-scale graphs and patterns, we propose a learning framework that augments different representation learning architectures and iteratively attends pattern and target data graphs to memorize \revisexin{intermediate states of subgraph isomorphism searching} for global counting.
We develop both {small} graphs ($\leq$ 1,024 subgraph isomorphisms in each) and {large} graphs ($\leq$  4,096 subgraph isomorphisms in each) sets to evaluate different \revisexin{representation and interaction modules}.
\revisexin{A mutagenic compound dataset, \textit{MUTAG}, is also used to evaluate neural models and demonstrate the success of transfer learning.}
While the learning based approach is inexact, we are able to generalize to count large patterns and data graphs in linear time compared to the exponential time of the original NP-complete problem.
Experimental results show that learning based subgraph isomorphism counting can speed up the traditional algorithm, VF2, \revisexin{10-\reviseyq{1,000}} times
with acceptable errors.
Domain adaptation based on fine-tuning also \reviseyq{shows} the usefulness of our approach in real-world applications.
\end{abstract}




\maketitle

\section{Introduction}


Counting subgraph isomorphisms aims at determining the number of subgraphs of a given graph that match (i.e., be isomorphic to) a specific \textit{pattern} graph of interest. 
It is one of the useful graph-related tasks for knowledge discovery and data mining applications, 
e.g., discovering protein interactions in bioinformatics~\citep{alon-network-motifs,AlonDHHS08}, finding active chemical compounds for new drug development in chemoinformatics~\citep{HuanWP03}, and mining social structures in online social network analysis~\citep{KuramochiK04}.
Moreover, finding sub-structures are useful for  heterogeneous information networks (HINs) such as knowledge graphs and recommender systems~\cite{FangLZWCL16,HuangZCSML16,HeJiang2017,HuanZhao2017}.
The rich types on HIN's schema provide meaningful semantics so that counting itself is valuable for a variety types of queries. 
For example, in a knowledge graph, we can answer questions like ``how many languages are there in Africa spoken by people living near banks of the Nile River?'' On a social network, a query like 
``number of pairs of friends, among them one likes a comic, and the other one likes its movie adaptation'' 
may help briefing an advertisement targeting.

However, previous studies on similar tasks are often under specific constraints, and discussions for general patterns on large heterogeneous graphs are limited. 
Particularly, former studies on counting reduce to small subgraphs with 3-5 nodes as \textit{motifs} (without considering isomorphisms)~\cite{alon-network-motifs} or \textit{graphlets} (limited to induced subgraphs)~\cite{PrzuljCJ06}.
Moreover, although customized pattern mining algorithms or graph database indexing based approaches may tackle the general problem of subgraph isomorphism~\citep{ullmann1976an, vento2004a, HeS08, Han2013turbo, CarlettiFSV18}, all such algorithms struggle with scaling up to larger graphs due to the NP-complete nature of the task. 
To extend to larger graphs under the more general and theoretically fundamental setting, alternative scalable methods are required. Specifically, we turn to approximate the answer by machine learning.


Learning with graphs has recently drawn much attention as neural approaches to representation learning have been proven to be effective for complex data structures~\citep{NiepertAK16,kipf2017semi,HamiltonYL17,schlichtkrull2018modeling,velivckovic2018graph,XuHLJ19}.
Admittedly, most of existing graph representation learning algorithms focus on tasks like node classification and link prediction~\citep{HamiltonYL17Survey} that information from {\it local} structure is enough for making inference.
Such tasks are different from ours which is to count over the whole graph.
However, recently it has been demonstrated that, more challenging reasoning problems that involve {\it global} inference, such as summary statistics, relational argmax, dynamic programming, and the NP-hard subset sum, can be elegantly solved by learning with properly designed architectures~\cite{XuLZDKJ20}.
The results are inspiring and encouraging. However, there are still challenges lying ahead to the target to count subgraph isomorphisms with machine learning:

$\bullet$ Typical sizes of existing datasets are limited,\footnote{\url{http://graphkernels.cs.tu-dortmund.de} provides several benchmark graph datasets of hundreds of small graphs; \url{http://networkrepository.com/} collects more graph data but most of them are still small.} and 
far from enough to train a graph neural network.
Meanwhile, it is non-trivial to generate sufficient training data for this problem due to the difficulty to obtain \reviseyq{the} ground truth, compared with other graph learning problems such as shortest path~\cite{graves2016hybrid} or centrality~\cite{FanZDCSL19}.

$\bullet$ Appropriate inductive bias must be carefully selected for such challenging task~\cite{abs-1806-01261}. Therefore, following considerations should be taken into account: 
First, as the general case of heterogeneous node types and edge types being considered, 
to represent and discriminate nodes and edges with neighborhood structure,
stronger and more complicated representation learning techniques should be applied.
Second, the task is not performed on a single graph but involves with an extra pattern. Therefore, the interaction between two graphs must be taken into consideration in the model design.
Third, the model would be trained on and applied to dataset with not only large graphs but also large number of samples, given the diversity of possible patterns. Hence the computational efficiency should be considered as well.




In this paper, to tackle the challenges above,  we first develop two large-scale datasets for our experiments including one with {\it small} graphs ($\leq$ 1,024 counts in each graph)
and one with {\it large} graphs ($\leq$ 4,096 counts in each graph).
We develop a framework to learn from our datasets.
To find appropriate inductive biases in graph neural networks, we compare different methods for extracting representations of node and edges, including CNN~\citep{kim2014convolutional}, RNN like GRU~\citep{ChoMGBBSB14}, attention models like Transformer-XL~\citep{transformerxl2019dai},  RGCN~\citep{schlichtkrull2018modeling}, a natural generalization of graph convolutional networks for heterogeneous graphs, \reviseyq{and RGIN, an extension of RGCN with the recent graph isomorphism network (GIN)~\cite{XuHLJ19}}.
Furthermore, we develop a dynamic intermedium attention memory network (DIAMNet) which iteratively attends the query pattern and the target data graph to scan the data graph with local structure memorized for global inference.
Then we perform systematic evaluation on our synthetic datasets to demonstrate the efficacy of inductive biases we have introduce. 
In addition, we also conduct experiments on the well-known benchmark dataset, \textit{MUTAG}, to show the power and usefulness of our approach for real applications.




Our main contributions are summarized as follows:

$\bullet$ ~To our best knowledge, this is the first work to model the subgraph isomorphism counting problem as a learning task, for which both training and inference enjoy linear time complexities, scalable for pattern/graph sizes and the number of data examples.

$\bullet$ ~We exploit the representation power of different deep neural network architectures under an end-to-end learning framework. In particular, we provide universal encoding methods for both sequence models and graph models, and upon them we introduce a dynamic intermedium attention memory network to address the more global inference problem of counting. 

$\bullet$ ~We conduct extensive experiments on two synthetic datasets and a well-known real benchmark dataset, which demonstrate that our framework can achieve good results on both relatively large graphs and patterns.
\revisexin{Data and code are available at \\ \url{https://github.com/HKUST-KnowComp/NeuralSubgraphCounting}.}

\section{Preliminaries}

We begin by introducing \reviseyq{the problem and our general idea.}
\subsection{Problem Definition}
\label{sec:problem_definition}
Traditionally, the subgraph isomorphism problem is defined between two simple graphs or two directed simple graphs, which is an NP-complete problem. 
We extend the problem to the more general case on directed heterogeneous multigraphs.

A graph or a pattern is defined as $\mathcal{G}=(\mathcal{V}, \mathcal{E}, \mathcal{X}, \mathcal{Y})$ where 
$\mathcal{V}$ is the set of vertices, each with a different \textit{vertex id}. $\mathcal{E} \subseteq \mathcal{V} \times \mathcal{V}$ is the set of edges,
$\mathcal{X}$ is a label function that maps a vertex to a \textit{vertex label}, and $\mathcal{Y}$ is a label function that maps an edge to a set of \textit{edge labels}.
Under the settings of heterogeneous multigraph, each multiedge can be represented by an ordered pair of source and target, along with a set of edge labels. That is to say, each edge can be uniquely identified with source, target, and its edge label.
To simplify the statement, we assume $\mathcal{Y}((u, v)) = \phi$ if $(u, v) \not \in \mathcal{E}$.
Here, we focus on isomorphic mappings that preserve graph topology, vertex labels and edge labels, but not vertex ids. More precisely, a pattern $\mathcal{G}_P=(\mathcal{V}_P,\mathcal{E}_P, \mathcal{X}_P, \mathcal{Y}_P)$ is \textit{isomorphic} to a graph $\mathcal{G}_G=(\mathcal{V}_G,\mathcal{E}_G, \mathcal{X}_G, \mathcal{Y}_G)$ if
there is a bijection $f: \mathcal{V}_G \rightarrow \mathcal{V}_P$ such that:
\begin{itemize}
    \item $\forall v \in \mathcal{V}_G, \mathcal{X}_G(v) = \mathcal{X}_P(f(v))$, 
    \item $\forall v \in \mathcal{V}_P, \mathcal{X}_P(v) = \mathcal{X}_G(f^{-1}(v))$,
    \item $\forall (u, v) \in \mathcal{E}_G, \mathcal{Y}_G((u, v)) = \mathcal{Y}_P((f(u), f(v)))$,
    \item $\forall (u, v) \in \mathcal{E}_P, \mathcal{Y}_P((u, v)) = \mathcal{Y}_G((f^{-1}(u), f^{-1}(v)))$.
\end{itemize}
Then,  a pattern $\mathcal{G}_P$ being isomorphic to a graph $\mathcal{G}_G$ is denoted as $\mathcal{G}_P \simeq \mathcal{G}_G$ and the function $f$ is named as an \textit{isomorphism}.
Moreover, a pattern $\mathcal{G}_P=(\mathcal{V}_P,\mathcal{E}_P, \mathcal{X}_P, \mathcal{Y}_P)$ is \textit{isomorphic to a subgraph} $\mathcal{G}_G'=(\mathcal{V}_G',\mathcal{E}_G', \mathcal{X}_G, \mathcal{Y}_G)$ of a graph $\mathcal{G}_G=(\mathcal{V}_G,\mathcal{E}_G, \mathcal{X}_G, \mathcal{Y}_G)$: 
$\mathcal{V}_G' \subseteq \mathcal{V}_G$, $\mathcal{E}_G' \subseteq \mathcal{E}_G \cap (\mathcal{V}_G' \times \mathcal{V}_G')$ if $\mathcal{G}_P \simeq \mathcal{G}_G'$.
The bijection function $f: \mathcal{V}_G' \rightarrow \mathcal{V}_P$ is named as a \textit{subgraph isomorphism}.
 
\begin{figure}
\centering
\includegraphics[width=1\linewidth]{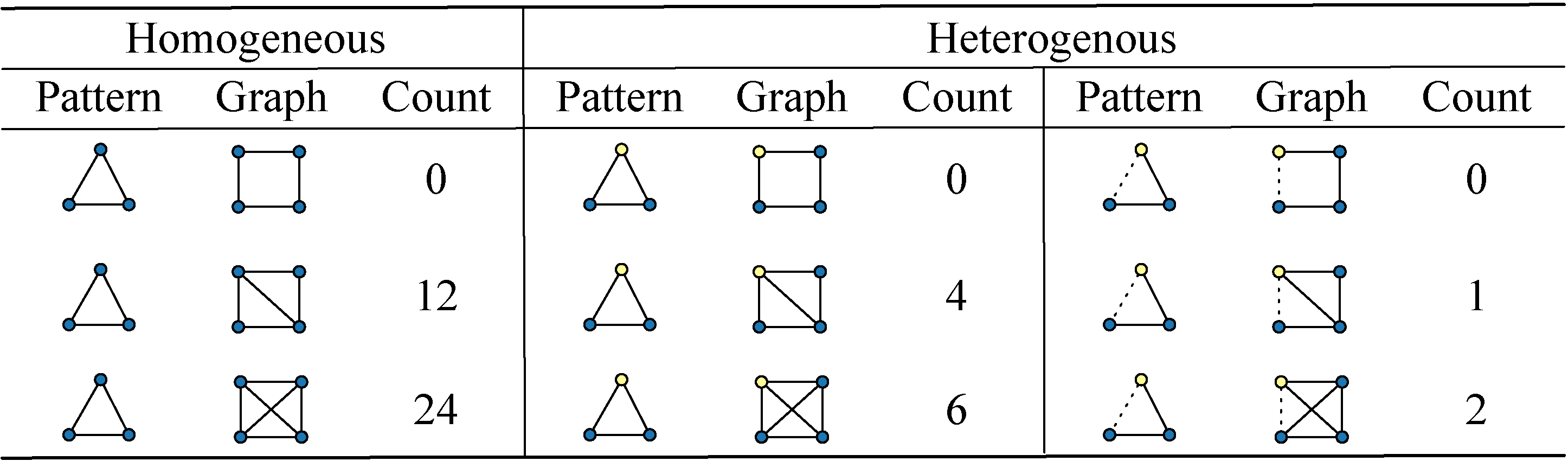}
\vspace{-0.2in}
\caption{Subgraph isomorphism counting examples in different settings: homogeneous graphs, heterogeneous vertex graphs where there are two types of nodes, and heterogeneous vertex and heterogeneous edge graphs where there are two types of nodes and two types of edges. }
\vspace{-0.2in}
\label{fig:prob-def}
\end{figure}

The subgraph isomorphism counting problem is defined as to find the number of all different subgraph isomorphisms between a pattern graph $\mathcal{G}_P$ and a graph $\mathcal{G}_G$.
Examples are shown in Figure~\ref{fig:prob-def}.
    
\subsection{General Idea}
Intuitively, we need to compute $\mathcal{O}(Perm(|\mathcal{V}_G|, |\mathcal{V}_P|) \cdot d^{|\mathcal{V}_P|})$ to solve the subgraph isomorphism counting problem by brute force, where $Perm(n, k) = \frac{n!}{(n-k)!}$,
$|\mathcal{V}_G|$ is the number of graph nodes, and $|\mathcal{V}_P|$ is the number of pattern nodes, $d$ is the maximum degree.
Ullmann's algorithm~\citep{ullmann1976an}, the standard exact method on the task, reduces the seaching time to $\mathcal{O}({|\mathcal{V}_P|}^{|\mathcal{V}_G|} \cdot {|\mathcal{V}_G|}^2)$.

Since the computational cost grows exponentially, it is impossible to be applied to larger graphs or patterns. As for non-exponential approximations, we may apply neural networks to learn distributed representations for $\mathcal{G}_G$ and $\mathcal{G}_P$ with a linear cost.
Then, to make predictions, pairwise interactions between nodes or edges of both the graph and the pattern can be handled by attention~\cite{bahdanau2014neural}, under the complexity of $\mathcal{O}(|\mathcal{V}_P| \cdot |\mathcal{V}_G| + |\mathcal{V}_P|^2 + |\mathcal{V}_G|^2)$ or $\mathcal{O}(|\mathcal{E}_P| \cdot |\mathcal{E}_G| + |\mathcal{E}_P|^2 + |\mathcal{E}_G|^2)$.
However, such quadratic complexity is still unacceptable when querying over large graphs. 
Unfortunately, if the attention module (especially the self-attention module) for handling interactions is completely absent, even though the time can be reduced to $\mathcal{O}(|\mathcal{V}_G|)$ or $\mathcal{O}(|\mathcal{E}_G|)$, the capability of the model would be strictly limited and the performance may deteriorate. Therefore, in this work we introduce the attention mechanism with an additional memory to reduce the complexity to approximately linear, while we expect the performance to be kept acceptable.
\section{Methodologies}

\begin{figure}[t]
\centering
\includegraphics[width=1.0\linewidth]{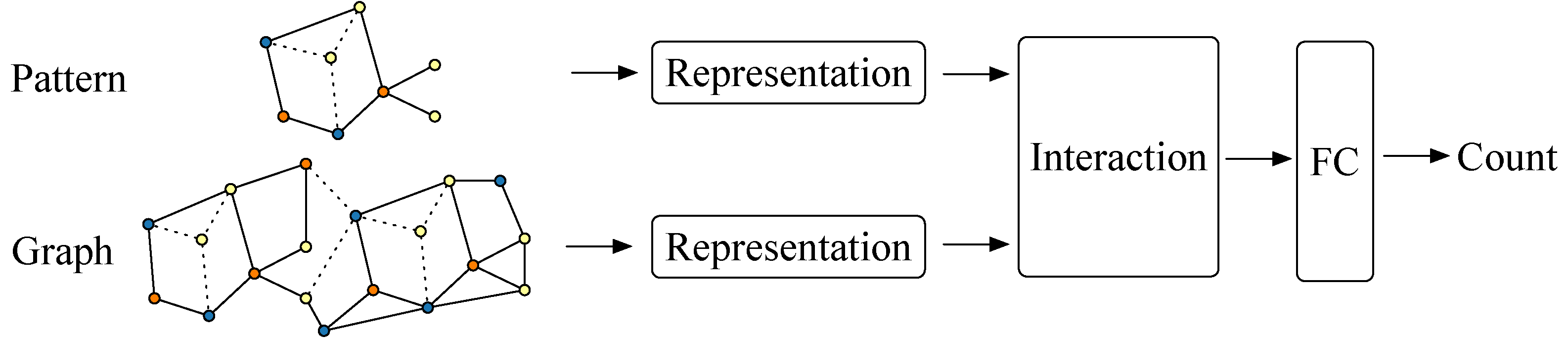}
\vspace{-0.2in}
\caption{General framework of neural subgraph isomorphism counting models.}
\vspace{-0.1in}
\label{fig:general-framework}
\end{figure}

\begin{figure}[t]
\centering
\vspace{-0.2in}
\includegraphics[width=1.0\linewidth]{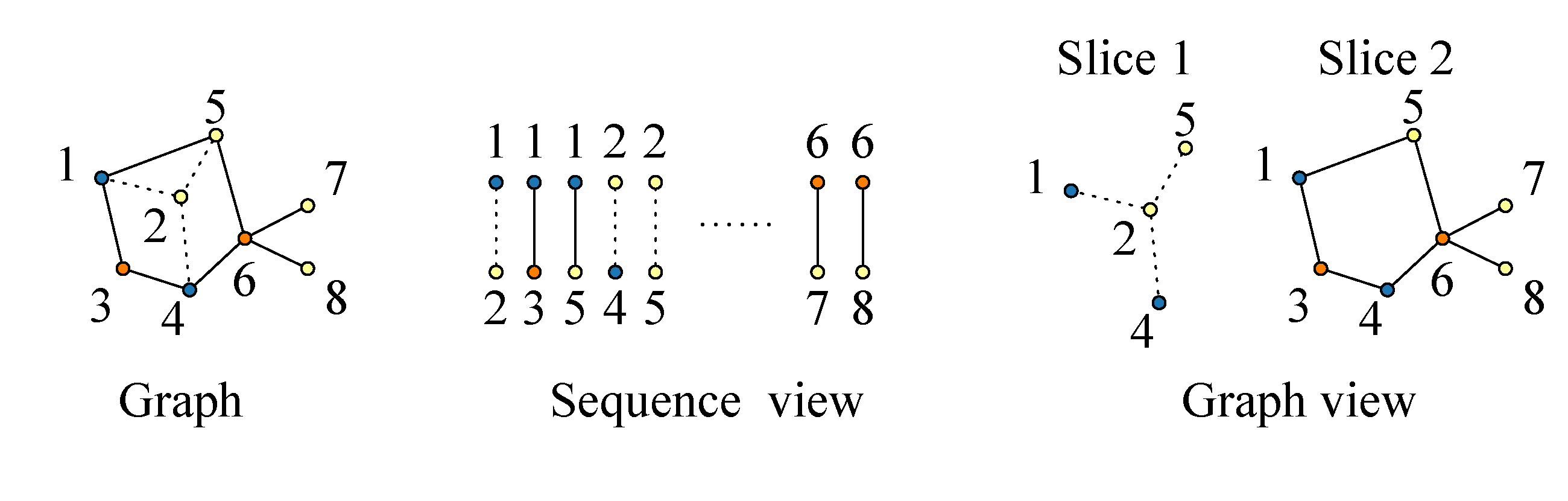}
\vspace{-0.4in}
\caption{Two views of different encoding methods. There are three labels of vertices (marked as orange, blue, and yellow) and two labels of edges (marked as solid and dashed lines). The numbers affixed show one possible set of ids for the vertices.}
\vspace{-0.2in}
\label{fig:two-repr}
\end{figure}

A general framework of learning to count subgraph isomorphism is shown in Figure \ref{fig:general-framework}.
In our framework, we treat the graph $\mathcal{G}_G$ as the data containing information and the pattern $\mathcal{G}_P$ as the query to retrieve information from the graph.
Thus the counting problem can be naturally formulated as a question-answering problem well established in natural language processing (e.g., \cite{kumar2016ask}).
Besides the difference of text and graph as inputs, the major problem is that in a QA model, the answers are usually facts extracted from the provided texts while in the counting problem, the answers are summary statistics of matched local patterns.
To make the problem learnable, the graph (or the pattern) should be either represented as a sequence of edges, or a series of adjacent matrices and vertex features. 
The difference between sequence encoding and graph encoding is shown in Figure~\ref{fig:two-repr}.
For sequence inputs 
we can use CNNs~\citep{kim2014convolutional}, RNNs such as LSTM~\citep{HochreiterS97}, or Transformer-XL~\citep{transformerxl2019dai} to extract high-level features.
If the inputs are modeled as series of adjacent matrices and vertex features, 
we can use RGCN~\citep{schlichtkrull2018modeling} \revisexin{or an extension of GIN~\cite{XuHLJ19}} to learn vertex representations with message passing from neighborhoods. 
After obtaining the pattern representations and the graph representations, we feed them into an interaction module to extract the correlated features from each side. Then we feed the output context of the interaction module with size information into fully-connected layers to make predictions.

\subsection{Sequence Models}
\subsubsection{Sequence Encoding}
\label{sec:sequence_encoding}

In sequence models, the minimal element of a graph (or a pattern) is an edge.
By definition, at least three attributes are required to identify an edge $e$, 
which are the source vertex id $u$, the target vertex id $v$, and its edge label $y \in \mathcal{Y}(e)$.
We further add two attributes of vertices' labels to form a 5-tuple $(u, v, \mathcal{X}(u), y, \mathcal{X}(v))$ to represent an edge $e$, 
where $\mathcal{X}(u)$ is the source vertex label and $\mathcal{X}(v)$ is the target vertex label.
A list of 5-tuple is referred as a code. We follow the order defined in gSpan~\citep{yan2002gspan} to compare pairs of code lexicographically; the detailed definition is given in Appendix~\ref{appendix:lexicographic_order}.
The minimum code is the code with the minimum lexicographic order with the same elements. 
Finally, each graph can be represented by the corresponding minimum code, and vice versa.

Given that a graph is represented as a minimum code, or a list of 5-tuples, the next encoding step is to transform each 5-tuple into a vector.
Assuming that we know the max values of $|\mathcal{V}|$, $|\mathcal{X}|$, $|\mathcal{Y}|$ in a dataset in advance, 
we can encode each vertex id, vertex label, and edge label into $B$-nary digits, where $B$ is the base and each digit $d \in \{0, 1, \cdots, B-1\}$.
We can represent each digit as a one-hot vector so that each id or label can be vectorized as a multi-hot vector which is the concatenation of these one-hot vectors.
Furthermore, the 5 multi-hot vectors can be concatenated to represent a 5-tuple.
The length of the multi-hot vector of a 5-tuple is $B \cdot (2 \cdot \lceil \log_{B}(Max(|\mathcal{V}|)) \rceil + 2 \cdot \lceil \log_{B}(Max(|\mathcal{X}|)) \rceil + \lceil \log_{B}(Max(|\mathcal{Y}|)) \rceil)$. 
Then minimum is achieved when $B=2$.
\revisexin{We can easily calculate the dimension $d_g$ for graph tuples and the dimension $d_p$ for pattern tuples.}
Furthermore, the minimum code can be encoded into a multi-hot matrix, $\bm{G} \in \mathbb{R}^{|\mathcal{E}_G| \times d_g}$ for a graph $\mathcal{G}_G$ or $\bm{P} \in \mathbb{R}^{|\mathcal{E}_P| \times d_p}$ for a pattern $\mathcal{G}_P$.

This encoding method can be extended when we have larger values of $|\mathcal{V}|$, $|\mathcal{X}|$, $|\mathcal{Y}|$.
A larger value, e.g., $|\mathcal{V}|$, only increases the length of its multi-hot vector corresponding to additional digits. Therefore, we can regard new digits as the same number of \reviseyq{zeroes and ones} \revisexin{ahead} the previous encoded vector, e.g., extending $0110$ to $\underline{01 \cdots 01}0110$.
\revisexin{As long as we initialize the model parameters associated with new extended dimensions as zeros, the extended input and parameters will not affect models.}

\subsubsection{Sequence Neural Networks}
Given the encoding method in Section~\ref{sec:sequence_encoding}, we can simply embed graphs as multi-hot matrices. Then we can use general strategies of sequence modeling to learn dependencies among edges in graphs.


\noindent\textbf{Convolutional Neural Networks (CNNs)} have been proved to be effective in sequence modeling \citep{kim2014convolutional}. 
In our experiments, we \revisexin{stack multiply convolutional layers and max-pooling layers} to obtain a sequence of high-level features. 

\noindent \revisexin{\textbf{Long Short-Term Memory (LSTM)}~\citep{HochreiterS97} is widely used in many sequence modeling tasks. Its memory cell is considered to be capable of learning long-term dependencies.}

\noindent \textbf{Transformer-XL (TXL)}~\citep{transformerxl2019dai} is a variant of the Transformer architecture~\citep{vaswani2017attention} and enables learning long dependencies beyond a fixed length without disrupting temporal coherence. \revisexin{Unlike the original autoregressive settings, we use the Transformer-XL as a feature extractor, in which the attention mechanism has a full, unmasked scope over the whole sequence.}
But its computational cost grows quadratically with the segment length and the memory size, so the tradeoff between performance and efficiency should be considered.

\subsection{Graph Models}
\subsubsection{Graph Encoding}
\label{sec:graph_encoding}


In graph models, each vertex has a feature vector, and each edge is used to pass information from its source to its target.
GNNs do not need vertex ids and edge ids explicitly because the adjacency information is included in an adjacent matrix.
As explained in Section~\ref{sec:sequence_encoding}, we can vectorize vertex labels into multi-hot vectors as vertex features.
\revisexin{The adjacent information of a (directed or undirected) simple graph can be stored in a sparse matrix to reduce the memory usage and improve the computation speed.}
As for heterogeneous graphs, behaviors of edges should depend on edge labels.
\revisexin{Therefore, relation-specific transformations to mix the edge label and topological information \reviseyq{should be applied and passed} to the target.}


\subsubsection{Graph Neural Networks}
We set encoded vertex labels as vertex features and propagate mixed messages by edges.

\noindent \textbf{Relational Graph Convolutional Network (RGCN)}~\citep{schlichtkrull2018modeling} is developed specifically to handle multi-relational data in realistic knowledge bases. Each relation corresponds to a transformation matrix to transform relation-specific information from \revisexin{the source to the target}. 
Two decomposition methods are proposed to address the rapid growth in the number of parameters with the number of relations: basis-decomposition and block-diagonal-decomposition. 
\revisexin{We choose the block-diagonal-decomposition and follow the original setting with the mean aggregator.}

\noindent \textbf{Graph Isomorphism Network (GIN)}~\cite{XuHLJ19} can capture homogeneous graph structures better with the help of Multilayer Perceptrons (MLPs) and the sum aggregator. We add MLPs after each relational graph convolutional layer and use the sum aggregator to aggregate messages, and we name this variant as \textbf{RGIN}.

\begin{figure}
\vspace{-0.1in}
\centering
\includegraphics[width=1\linewidth]{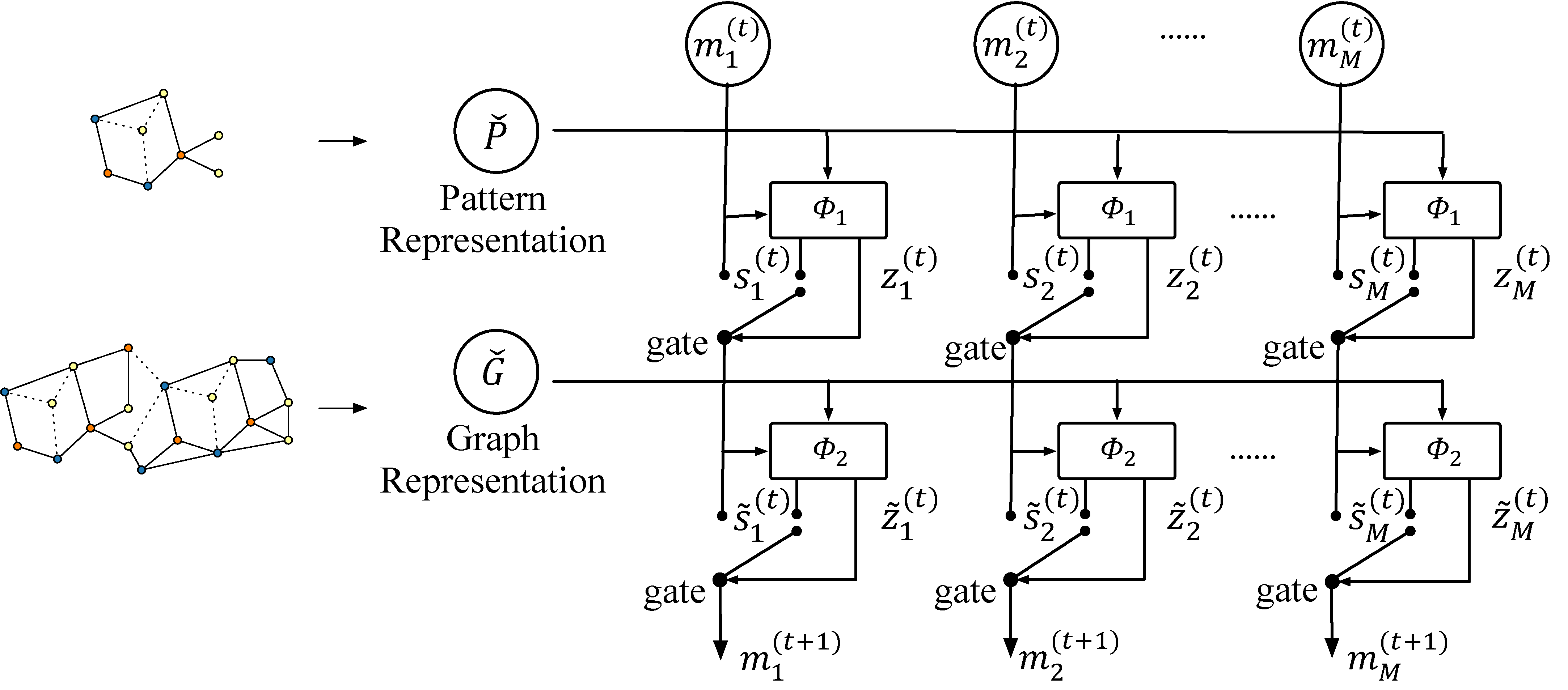}
\vspace{-0.2in}
\caption{Dynamic intermedium attention memory network (DIAMNet). $\Phi_1$ represents Eqs. (\ref{eq:phi11}) and  (\ref{eq:phi12}), $\Phi_2$ represents Eqs. (\ref{eq:phi21}) and (\ref{eq:phi22}), and two types of gates are Eqs. (\ref{eq:gate1}) and  (\ref{eq:gate2}).}
\label{fig:memnet}
\vspace{-0.2in}
\end{figure}

\subsection{Dynamic Intermedium Attention Memory}
\label{sec: attn_mem}

After obtaining a graph representation $\check{\bm{G}}$ and a pattern representation $\check{\bm{P}}$ from a sequence model or a graph model where their column vectors are $d$-dimensional, we feed them as inputs of interaction layers to extract the correlated context between the pattern and the graph. A naive idea is to use attention mechanism~\citep{bahdanau2014neural} to model interactions between these two representations and interactions over the graph itself. 
However, this method is not practical due to its complexity, up to $\mathcal{O}(|\mathcal{E}_P| \cdot |\mathcal{E}_G| + |\mathcal{E}_G|^2)$ for sequence modeling and  $\mathcal{O}(|\mathcal{V}_P| \cdot |\mathcal{V}_G| + |\mathcal{V}_G|^2)$ for graph modeling.

To address the problem of high computational cost in the attention mechanism, we propose the Dynamic Intermedium Attention Memory Network (DIAMNet), using an external memory as an intermedium to attend both the pattern and the graph in order. To make sure that the memory has the knowledge of the pattern while attending the graph and vice-versa, this dynamic memory is designed as a gated recurrent network as shown in Figure~\ref{fig:memnet}. Assuming that the memory size is $M$ and we have $T$ recurrent steps, the time complexity is decreased to $\mathcal{O}(T \cdot M \cdot (|\mathcal{E}_P| + |\mathcal{E}_G|))$ or $\mathcal{O}(T \cdot M \cdot (|\mathcal{V}_P| + |\mathcal{V}_G|))$, which means the method can be easily applied to large-scale graphs.

The external memory $\bm{M}$ is divided into $M$ blocks $\{\bm{m}_1, ..., \bm{m}_M \}$, where $\bm{m}_j \in \mathbb{R}^{d}$. At each time step $t$, $\{\bm{m}_j\}$ is updated by the pattern and the graph in order via multi-head attention mechanism~\citep{vaswani2017attention}. Specifically, the update equations of our DIAMNet are given by:
{\footnotesize
\begin{align}
    \bm{s}_j^{(t)} &= MultiHead(\bm{m}_j^{(t)}, \check{\bm{P}}, \check{\bm{P}}) \label{eq:phi11},\\
    \bm{z}_j^{(t)} &= \sigma(\bm{U}_P \bm{m}_j^{(t)} + \bm{V}_P \bm{s}_j^{(t)}) \label{eq:phi12},\\
    \overline{\bm{s}}_j^{(t)} &= \bm{z}_j^{(t)} \odot \bm{m}_j^{(t)} + (1 - \bm{z}_j^{(t)}) \odot \bm{s}_j^{(t)}, \label{eq:gate1}\\
    \widetilde{\bm{s}}_j^{(t)} &= MultiHead(\overline{\bm{s}}_j^{(t)}, \check{\bm{G}}, \check{\bm{G}}), \label{eq:phi21}\\
    \widetilde{\bm{z}}_j^{(t)} &= \sigma(\bm{U}_G \overline{\bm{s}}_j^{(t)} + \bm{V}_G \widetilde{\bm{s}}_j^{(t)}), \label{eq:phi22}\\
    \bm{m}_j^{(t + 1)} &= \widetilde{\bm{z}}_j^{(t)} \odot \overline{\bm{s}}_j^{(t)} + (1 - \widetilde{\bm{z}}_j^{(t)}) \odot \widetilde{\bm{s}}_j^{(t)}\label{eq:gate2}.
\end{align}
}
Here $MultiHead$ is the attention method described in~\citep{vaswani2017attention}, $\sigma$ represents the logistic sigmoid function, $\overline{\bm{s}}_j$ is the intermediate state of the $j^{th}$ block of memory that summarizes information from the pattern, and $\widetilde{\bm{s}}$ for information from both the pattern and the graph. $\bm{z}_j$ and $\widetilde{\bm{z}}_j$ are two gates designed to control the updates on the states in the $j^{th}$ block. $\bm{U}_P, \bm{V}_P, \bm{U}_G, \bm{V}_G \in \mathbb{R}^{d \times d}$ are trainable parameters.
There are many possible ways to initialize $\bm{M}^{(0)}$, and we discuss the memory initialization in Appendix~\ref{appendix:meminit}.

\section{Experiments}

In this section, we report our major experimental results.
\subsection{Datasets}

\begin{table}[t]
    \vspace{-0.1in}
    \caption{Statistics of the datasets. ``\#'' means number, ``A.'' means average, ``M.'' means max.}
    \vspace{-0.1in}
    \label{table:datasets}
    \centering
    \footnotesize
    \begin{tabular}{l|c|c|c|c|c|c}
    \toprule
        & \#Training & \#Dev & \#Test & A.$(|\mathcal{V}_G|)$ & A.$(|\mathcal{E}_G|)$ & M.(Counts)\\
        \midrule
        Small & 358,512 & 44,814 & 44,814 & 32.6 & 76.3 & 1,024 \\
        Large & 316,224 & 39,528 & 39,528 & 240.0 & 560.0 & 4,096 \\
        MUTAG & 1,488 & 1,512 & 1,512 & 17.9 & 39.6 & 156 \\
    \bottomrule
    \end{tabular}
    \vspace{-0.2in}
\end{table}

\label{sec: dataset_generation}
In order to train and evaluate our neural models for the subgraph isomorphism counting problem, we need to generate enough graph-pattern data.

\paragraph{Synthetic data} We generate two synthetic datasets, \textit{small} and \textit{large}.
As there's no special constraint on the pattern, the pattern generator may produce any connected multigraph without identical edges, i.e., parallel edges with identical label.
In contrast, the ground truth numbers of subgraph isomorphisms must be tractable in synthetic data graphs given one pattern. 
\revisexin{
We follow the idea of neighborhood equivalence class (NEC) in TurboISO \citep{Han2013turbo} to control the necessary conditions of a subgraph isomorphism in the graph generation process. The detailed algorithms are shown in Appendix~\ref{appendix:graph_generator}. 
Our graph generator first generates multiple disconnected components, possibly with some subgraph isomorphisms.
Then the generator merges these components into a larger graph and ensures that there is no more subgraph isomorphism created during the \reviseyq{merging} process. 
}
Hence, the subgraph isomorphism search can be done over small components in parallel before merging into a large graph.
Using the pattern generator and the graph generator above, we can generate \reviseyq{many} patterns and graphs for neural models. 
We are interested in follow research questions: whether sequence models and graph models can perform well given limited data, whether their running time is acceptable, whether memory can help models make better predictions even faced with a NP-complete problem, \revisexin{and whether models pre-trained on synthetic data can be \reviseyq{transferred to} real-life data}. 
To evaluate different neural architectures and different prediction networks, we generate two datasets in different graph scales, with statistics reported in Table~\ref{table:datasets}.
There are 187 unique patterns, where 75 patterns belong to the \textit{small} dataset \reviseyq{and} 122 patterns belong to the \textit{large} dataset.
\revisexin{Graphs are generated randomly based on these patterns.}
The generation details are reported in Appendix~\ref{appendix:datasets}.

\paragraph{MUTAG} We also generate 24 patterns for 188 graphs on the \textit{MUTAG} dataset.
We use the same pattern generator to generate 24 different heterogenous patterns, and their structures are listed in Appendix~\ref{appendix:mutag}.
To make the dataset more challenging, graphs in training data, dev data, and test data are disjoint.


\subsection{Implementation Details}
Instead of directly feeding multi-hot encoding vectors into representation modules, \revisexin{we use two linear layer to separately transform pattern and graph multi-hot vectors to vectors with the predefined hidden size.}
To improve the efficiency and performance, we also add a filter network to filter out edges or vertices in a graph that are irrelevant to the given pattern before representation modules. The details of this filter network is shown in Section~\ref{appendix:filternet}. 

\subsubsection{Representation Models}
In our experiments, we implemented five different representation models:
(1) \textbf{CNN} is a 3-layer convolutional layers followed by max-pooling layers. The convolutional kernels are 2,3,4 respectively and strides are 1. The pooling kernels are 2,3,4 and strides are 1. 
(2) \textbf{LSTM} is a simple 3-layer LSTM model.
(3) \textbf{TXL} is a 6-layer Transformer encoder with additional memory. 
(4) \textbf{RGCN} is a 3-layer RGCN with the block-diagonal-decomposition. We follow the same setting in \reviseyq{the original} paper to use mean-pooling in the message propagation part.
(5) \textbf{RGIN} \revisexin{is a combination of GIN and RGCN: adding a 2-layer MLP after each relational convolutional layer and using the sum aggregator.}
\revisexin{Residual connections~\cite{HeZRS16} are added to limit the vanish gradient problem.}

\subsubsection{Interaction Networks}
After getting a graph representation $\check{\bm{G}}$ and a pattern representation $\check{\bm{P}}$ from the representation learning modules, we feed them into the following different types of interaction layers for comparison. 




\noindent \textbf{Pool}: A simple pooling operation is directly applied for the output of neural models for classification or regression.
We apply one of sum-pooling, mean-pooling, and max-pooling for $\check{\bm{G}}$ and $\check{\bm{P}}$ to obtain $\check{\bm{g}}$ and $\check{\bm{p}}$, and then send $Concate(\check{\bm{g}}, \check{\bm{p}}, \check{\bm{g}}-\check{\bm{p}}, \check{\bm{g}} \odot \check{\bm{p}})$ with the graph size and the pattern size into the next fully connected (FC) layers.


\noindent \textbf{MemAttn}: QA models always use source attention and self-attention to find answers in paragraphs~\cite{Hu2018Reinforced}. 
One drawback of attention is the computational cost. We consider to use memory for the key and the value.
Two gated multi-head attention is applied on $\check{\bm{G}}$ and $\check{\bm{P}}$, which is described in Appendix~\ref{appendix:memattn}.
Finally, we get the mean-pooled representations for the pattern and the graph and feed the same combination in Pool to the FC layers.

\noindent \textbf{DIAMNet}: We compare the performance and efficiency of our DIAMNet proposed in Section~\ref{sec: attn_mem} with above interaction networks.
Similarly, we have many initialization strategies in Appendix~\ref{appendix:meminit}, but we find mean-pooling is fast and performs well.
Finally, we feed the whole memory with size information into the next FC layers.
The main difference between DIAMNet and MemAttn is that the memory in DIAMNet is regarded as the query, initialized once and updated $t$ times, but MemAttn initializes memory $2t$ times, takes memory as the key and the value, and updates graph representations $\check{\bm{G}}$ $t$ times.

\subsection{Experimental Settings}
For fair comparison, we set embedding dimensions, dimensions of all representation models, and the numbers of filters all as 128.
The segment size and memory size in TXL are also 64 based on the computational complexity consideration.
The length of memory in DIAMNet and MemAttn is fixed to 4 for all experiments.
We use the grid search to find the best recurrent steps in MemAttn and DIAMNet.
We use the mean squared error (MSE) to train models and choose best models based on the results of dev sets.
\revisexin{The optimizer is AdamW~\cite{LoshchilovH19} with a base learning rate of 1e-3 for synthetic data or 1e-4 for real-life data and a weight decay coefficient of 1e-6.}
To avoid gradient explosion and overfitting, we add gradient clipping and dropout with a dropout rate 0.2.
We use $Leaky\_ReLU$ as activation functions in all modules.
\revisexin{We use the same representation module with shared parameters to produce representations for both patterns and graphs considering the limited number of patterns.}

For the \textit{small} and \textit{large} datasets, we first train models on the \textit{small} and further use curriculum learning~\citep{BengioLCW09} on the \textit{large}.
For the \textit{MUTAG} dataset, we reports results of the best representation model, RGIN, with different numbers of training data.
\revisexin{Transfer learning~\cite{pan2009a} from synthetic data to \textit{MUTAG} is also conducted.}

\revisexin{Training and evaluating of neural models were finished on NVIDIA V100 GPUs under PyTorch and DGL frameworks. VF2 were finished with 16-core (32 threads) Intel E5-2620v4 CPUs in parallel.}

\subsection{Evaluation Metrics}

As we model this subgraph isomorphism counting problem as a learning problem, we use common metrics in regression tasks, including the root mean square error (RMSE) and the mean absolute error (MAE).
\revisexin{In this task, negative predictions are meaningless so that those predictions are regarded as zeros.}
Two trivial baselines, \textbf{Zero} that always predicts 0 and \textbf{Avg} that always predicts the average count of training data, are also used in comparison.






\subsection{Results and Analysis on Synthetic Data}

We first report results for the \textit{small} dataset in Table \ref{table:small_results} and results for the \textit{large} dataset in Table \ref{table:large_results}.
Only the best Pool result is reported for each representation module.
In addition to the trivial all-zero and average baselines and other neural network learning based baselines, we are also curious about to what extent our neural models can be faster than traditional searching algorithms. Therefore, we also compare the running time \revisexin{(I/O time excluded)}.
\revisexin{Considering that we use NEC to generate graphs instead of random generation, we compare with VF2 algorithm~\citep{vento2004a} which does not involve NEC or similar heuristic rules (e.g., rules in VF3~\citep{CarlettiFSV18}) for fairness.}
From the experiments, we can draw following observations and conclusions.

\begin{table}[t]
    \vspace{-0.1in}
    \footnotesize
    \caption{Results of different models on the \textit{small} dataset. Time is evaluated on the whole test set.}
    \vspace{-0.1in}
    \label{table:small_results}
    \centering
    \begin{tabular}{rl|c|c|c}
    \toprule
        \multicolumn{2}{c|}{\multirow{2}{*}{Models}} & \multicolumn{3}{c}{Test} \\
        &  & RMSE & MAE & Time (sec) \\
        \midrule
        \parbox[t]{10mm}{\multirow{3}{*}{CNN}}
        & MaxPool      & 45.606 & 8.807 & 0.808 \\
        & MemAttn      & 30.773 & 6.066 & 0.907 \\
        & DIAMNet      & 30.257 & 5.776 & 1.198 \\
        \hline
        \parbox[t]{10mm}{\multirow{3}{*}{LSTM}}
        & SumPool      & 32.572 & 6.353 & 11.091 \\
        & MemAttn      & 33.497 & 7.601 & 11.623 \\
        & DIAMNet      & 32.009 & 6.013 & 11.295 \\
        \hline
        \parbox[t]{10mm}{\multirow{3}{*}{TXL}}
        & MeanPool     & 41.102 & 8.033 & 8.131 \\
        & MemAttn      & 41.213 & 7.770 & 13.531 \\
        & DIAMNet      & 37.897 & 7.271 & 13.828 \\
        \hline
        \parbox[t]{10mm}{\multirow{3}{*}{RGCN}}
        & SumPool      & 29.648 & 5.747 & 5.400 \\
        & MemAttn      & 25.950 & 4.973 & 7.890 \\
        & DIAMNet      & 26.489 & 4.781 & 8.435 \\
        \hline
        \parbox[t]{10mm}{\multirow{3}{*}{RGIN}}
        & SumPool      & 20.689 & 3.657 & 5.599 \\
        & MemAttn      & 22.067 & 4.049 & 8.061 \\
        & DIAMNet      & \bf 20.267 & \bf 3.632 & 8.084 \\
        \hline
        \parbox[t]{5mm}{Zero} & & 68.457 & 14.825 & - \\
        \parbox[t]{5mm}{Avg} & & 66.832 & 23.633 & - \\
        \hline
        \parbox[t]{5mm}{VF2} & & 0.0 & 0.0 & $\sim$110  \\
        \bottomrule
    \end{tabular}
    \vspace{-0.1in}
\end{table}

{\bf Comparison of different representation architectures.}
As shown in Table \ref{table:small_results}, in general, graph models outperform most of the sequence models.
CNN is the worst model for the graph isomorphism counting problem.
The code order does not consider the connectivity of adjacent vertices and relevant label information. Hence, convolutional operations and pooling operations cannot extract useful local information but may introduce some noise.
From results of the \textit{large} dataset, we can see that CNN is also worse than others.
In fact, we observe CNN always predicts 0 for large graphs.
\revisexin{RNN and transformers are widely used in sequence tasks, and LSTM and TXL are advanced variants with memory.}
\revisexin{
We note that LSTM models with different interaction networks are constantly better than TXL.
Although memory of LSTM is much smaller than that of TXL, memory of LSTM can somehow memorize all information that has been seen previously but that of TXL is just the representation of the previous segment.}
In our experiments, the segment size is 64 so that TXL can not learn the global information at a time.
The local structure information also misleads TXL, which is consistent with CNN.
A longer segment set for TXL may lead to better results, but it will require much more GPU memory and much longer time for training.
\revisexin{Graph models perform always better than sequence models. It is not surprising because these graph convolutional operations are designed based on the topology structure.}
RGIN is much better than RGCN and other sequence models, which shows that MLPs and the sum aggregator are good at vertex representation learning and structure modeling in this task.
The mean aggregator can model the distribution of neighbors but the distribution may misguide models~\cite{XuHLJ19}.

\begin{table}[t]
\vspace{-0.1in}
    \footnotesize
    \caption{Results of different models on the \textit{large} dataset. Time is evaluated on the whole test set.}
    \vspace{-0.1in}
    \label{table:large_results}
    \centering
    \begin{tabular}{rl|c|c|c}
        \toprule
        \multicolumn{2}{c|}{\multirow{2}{*}{Model}}
        & \multicolumn{3}{c}{Test} \\
        &  & RMSE & MAE & Time (sec) \\
        \hline
        \parbox[t]{10mm}{\multirow{3}{*}{CNN}}
        & MaxPool & 162.145 & 27.135 & 2.334 \\
        & MeanAttn & 172.568 & 24.096 & 2.514 \\
        & DIAMNet & 125.009 & 18.642 & 5.968 \\
        \hline
        \parbox[t]{10mm}{\multirow{3}{*}{LSTM}}
        & SumPool & 131.778 & 19.239 & 16.701 \\
        & MeanAttn & 130.147 & 20.997 & 22.025 \\
        & DIAMNet & 129.253 & 19.098 & 22.511 \\
        \hline
        \parbox[t]{10mm}{\multirow{3}{*}{TXL}}
        & MeanPool & 161.649 & 23.952 & 26.457 \\
        & MeanAttn & 162.355 & 24.690 & 35.169 \\
        & DIAMNet  & 153.588 & 23.635 & 35.169 \\
        \hline
        \parbox[t]{10mm}{\multirow{3}{*}{RGCN}}
        & SumPool & 129.200 & 18.851 & 7.639 \\
        & MeanAttn & 115.981 & 18.897 & 12.851 \\
        & DIAMNet & 110.667 & 16.760 & 13.548 \\
        \hline
        \parbox[t]{10mm}{\multirow{3}{*}{RGIN}}
        & SumPool & 108.183 & 14.938 & 7.633 \\
        & MeanAttn & 111.218 & 16.494 & 12.549 \\
        & DIAMNet & \bf 103.849 & \bf 14.236 & 12.798 \\
        \hline
        \parbox[t]{5mm}{Zero} & & 219.795 & 32.868 & - \\
        \parbox[t]{5mm}{Avg} & & 217.325 & 57.300 & - \\
        \hline
        \parbox[t]{5mm}{VF2} & & 0.0 & 0.0 & $\sim5\times10^3$ \\
        \bottomrule
    \end{tabular}
    \vspace{-0.1in}
\end{table}

\begin{figure*}
\vspace{-0.1in}
    \centering
    \begin{subfigure}[b]{0.22\linewidth}
        \includegraphics[width=\linewidth,height=1.5in]{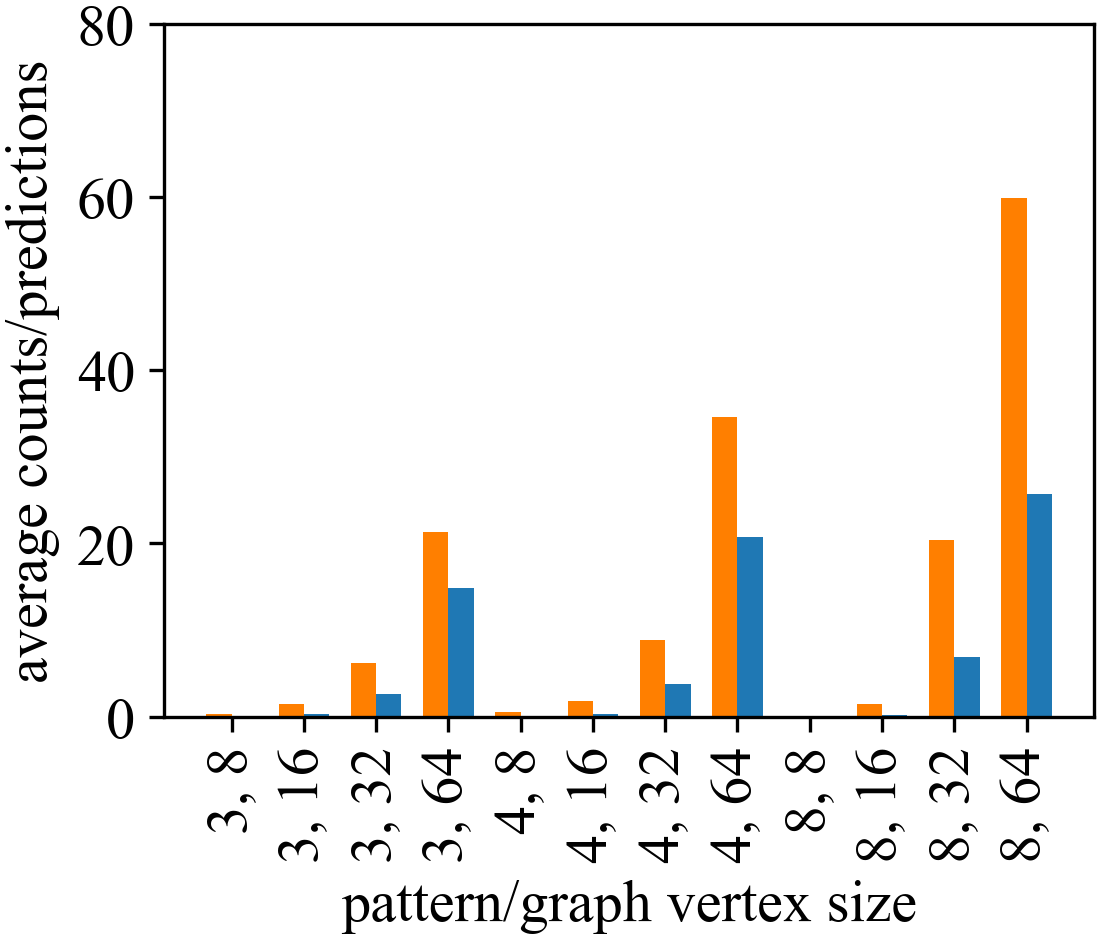}
        \caption{$O_\mathcal{V}$, MaxPool}
        \label{fig:CNN-Max-V-small}
    \end{subfigure}
    \begin{subfigure}[b]{0.22\linewidth}
        \includegraphics[width=\linewidth,height=1.5in]{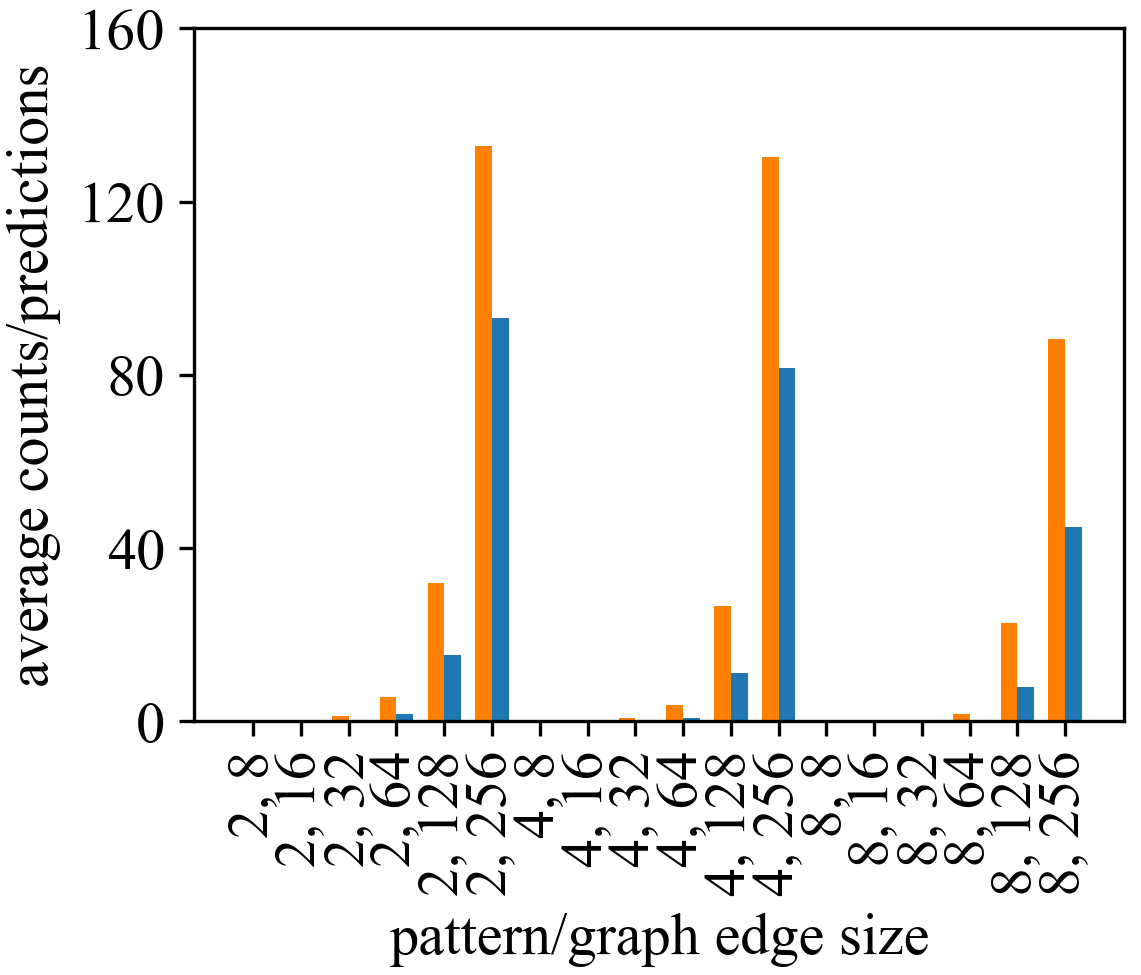}
        \caption{$O_\mathcal{E}$, MaxPool}
        \label{fig:CNN-Max-E-small}
    \end{subfigure}	
    \begin{subfigure}[b]{0.22\linewidth}
        \includegraphics[width=\linewidth,height=1.5in]{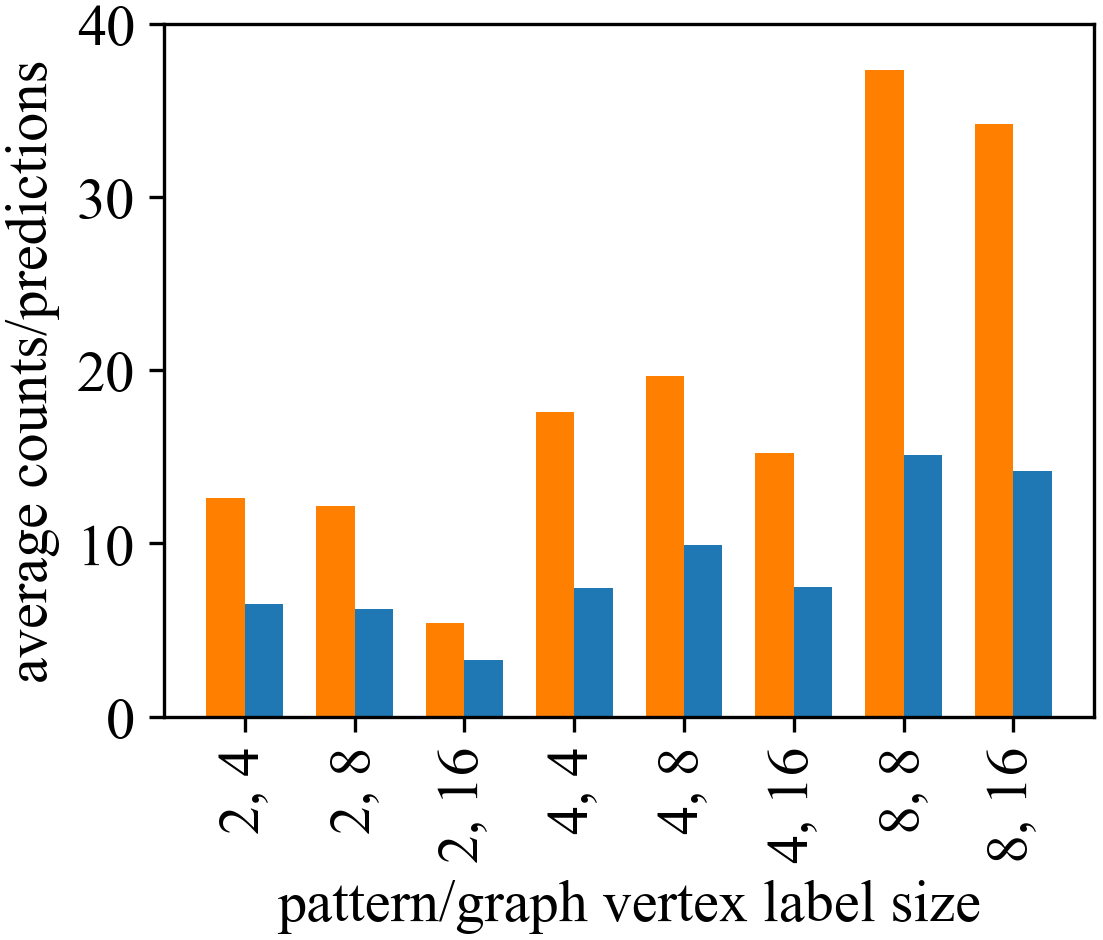}
        \caption{$O_\mathcal{X}$, MaxPool}
        \label{fig:CNN-Max-VL-small}
    \end{subfigure}	
    \begin{subfigure}[b]{0.22\linewidth}
        \includegraphics[width=\linewidth,height=1.5in]{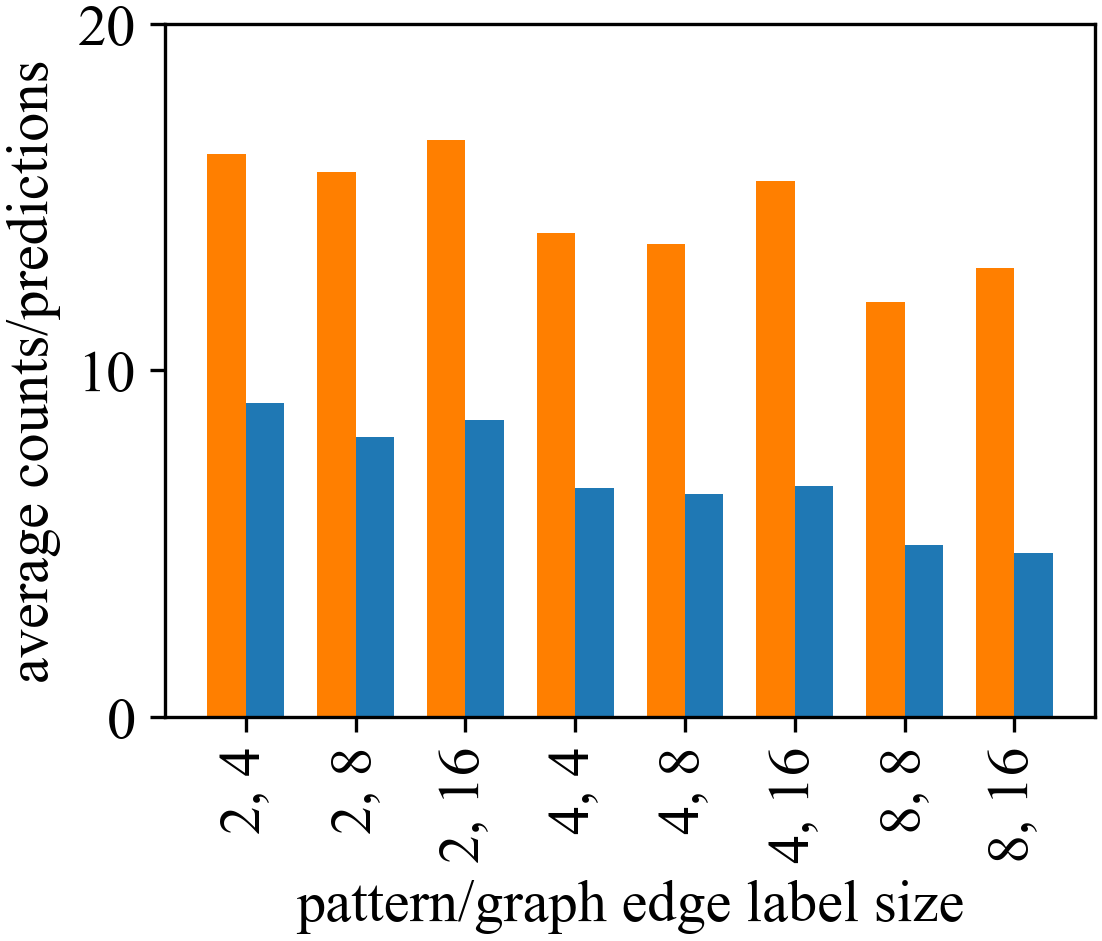}
        \caption{$O_\mathcal{Y}$, MaxPool}
        \label{fig:CNN-Max-EL-small}
    \end{subfigure}

	\begin{subfigure}[b]{0.22\linewidth}
        \includegraphics[width=\linewidth,height=1.5in]{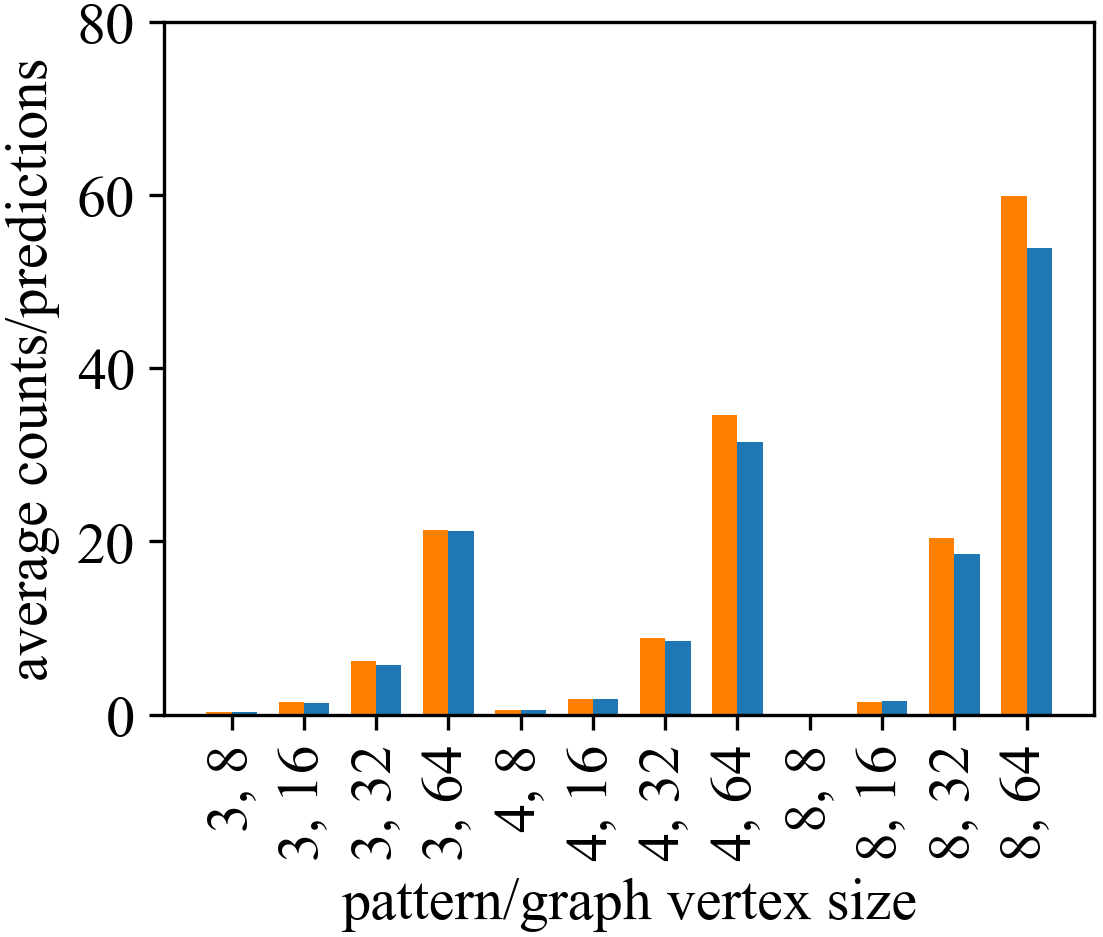}
        \caption{$O_\mathcal{V}$, DIAMNet}
        \label{fig:CNN-DIAMNet-V-small}
    \end{subfigure}
    \begin{subfigure}[b]{0.22\linewidth}
        \includegraphics[width=\linewidth,height=1.5in]{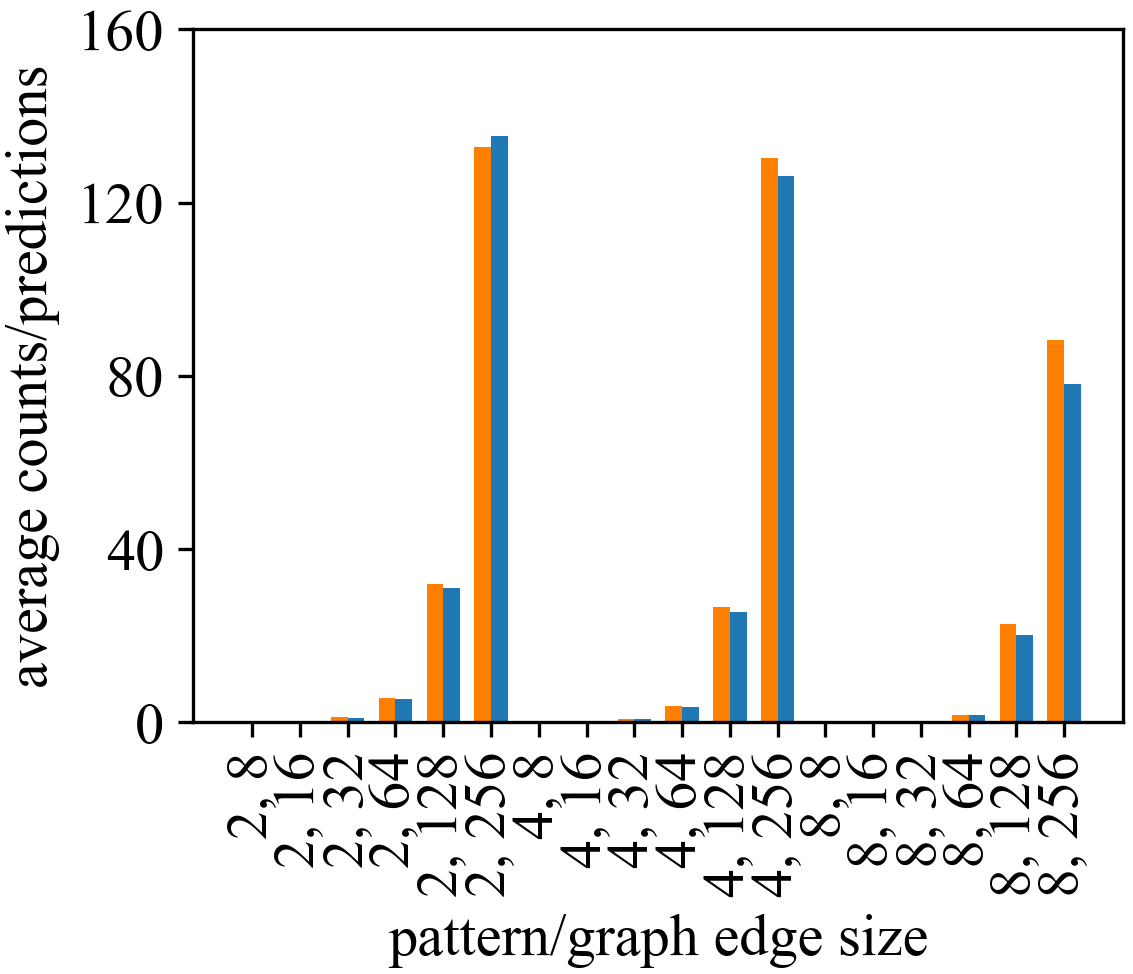}
        \caption{$O_\mathcal{E}$, DIAMNet}
        \label{fig:CNN-DIAMNet-E-small}
    \end{subfigure}	
    \begin{subfigure}[b]{0.22\linewidth}
        \includegraphics[width=\linewidth,height=1.5in]{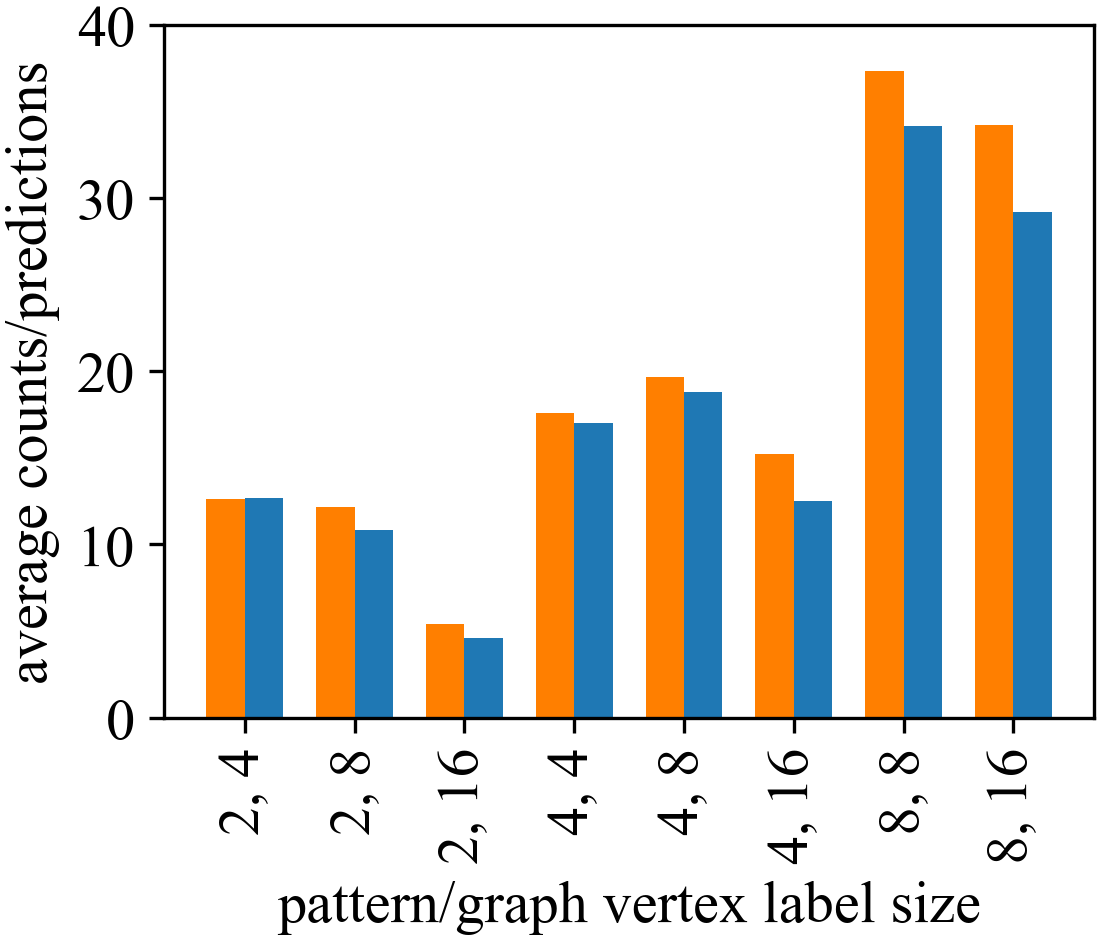}
        \caption{$O_\mathcal{X}$, DIAMNet}
        \label{fig:CNN-DIAMNet-VL-small}
    \end{subfigure}	
    \begin{subfigure}[b]{0.22\linewidth}
        \includegraphics[width=\linewidth,height=1.5in]{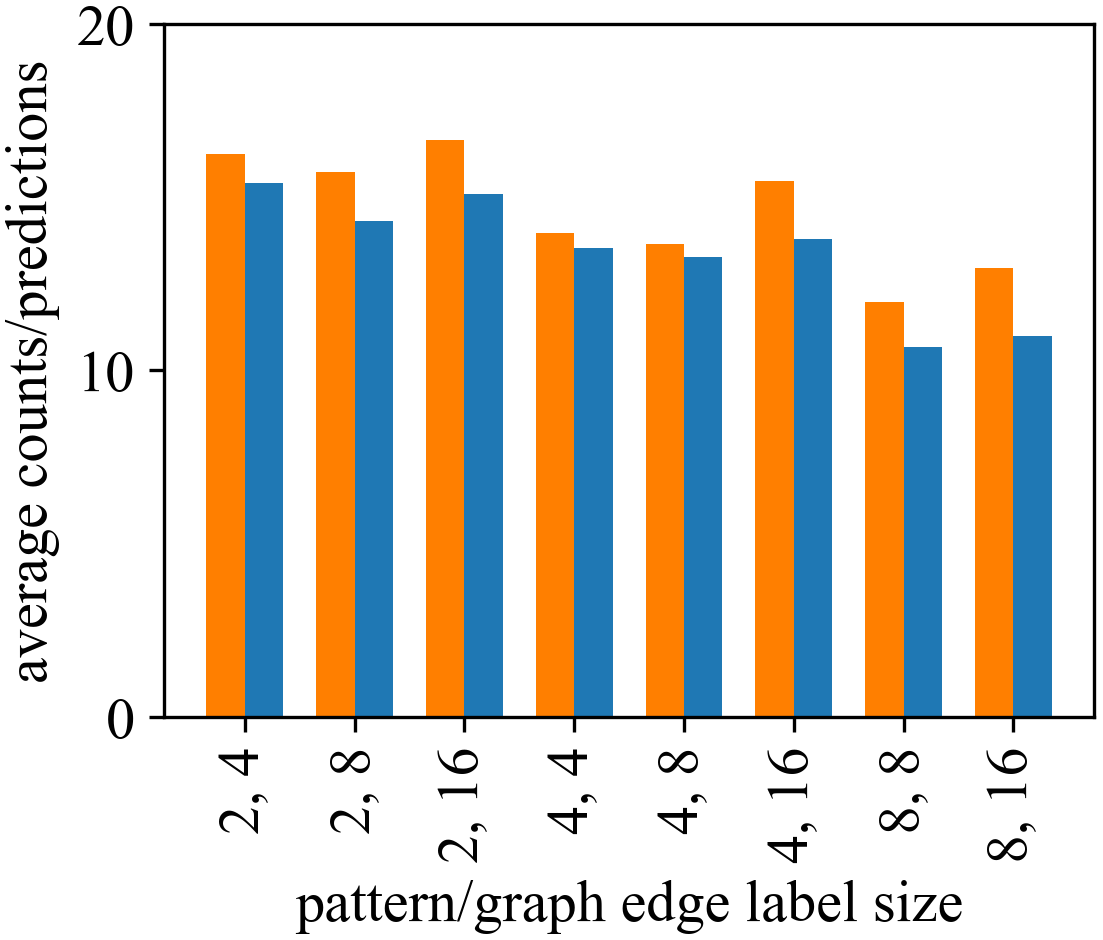}
        \caption{$O_\mathcal{Y}$, DIAMNet}
        \label{fig:CNN-DIAMNet-EL-small}
    \end{subfigure}
	\vspace{-0.1in}
    \caption{Model behaviors of CNN with MaxPool and CNN with DIAMNet on the \textit{small} dataset. Here $O_\mathcal{V}$ means that the bins on the x-axis are ordered by the size of pattern/graph vertex, $O_\mathcal{E}$ by the size of edge, $O_\mathcal{X}$ by the size of vertex label, and $O_\mathcal{Y}$ by the size of edge label. The orange color refers to the ground truth and the blue color refers to the predictions.}\label{fig:model-behavior-small}
    \vspace{-0.1in}
\end{figure*}

\begin{figure*}
    \centering
    \begin{subfigure}[b]{0.22\linewidth}
        \includegraphics[width=\linewidth,height=1.5in]{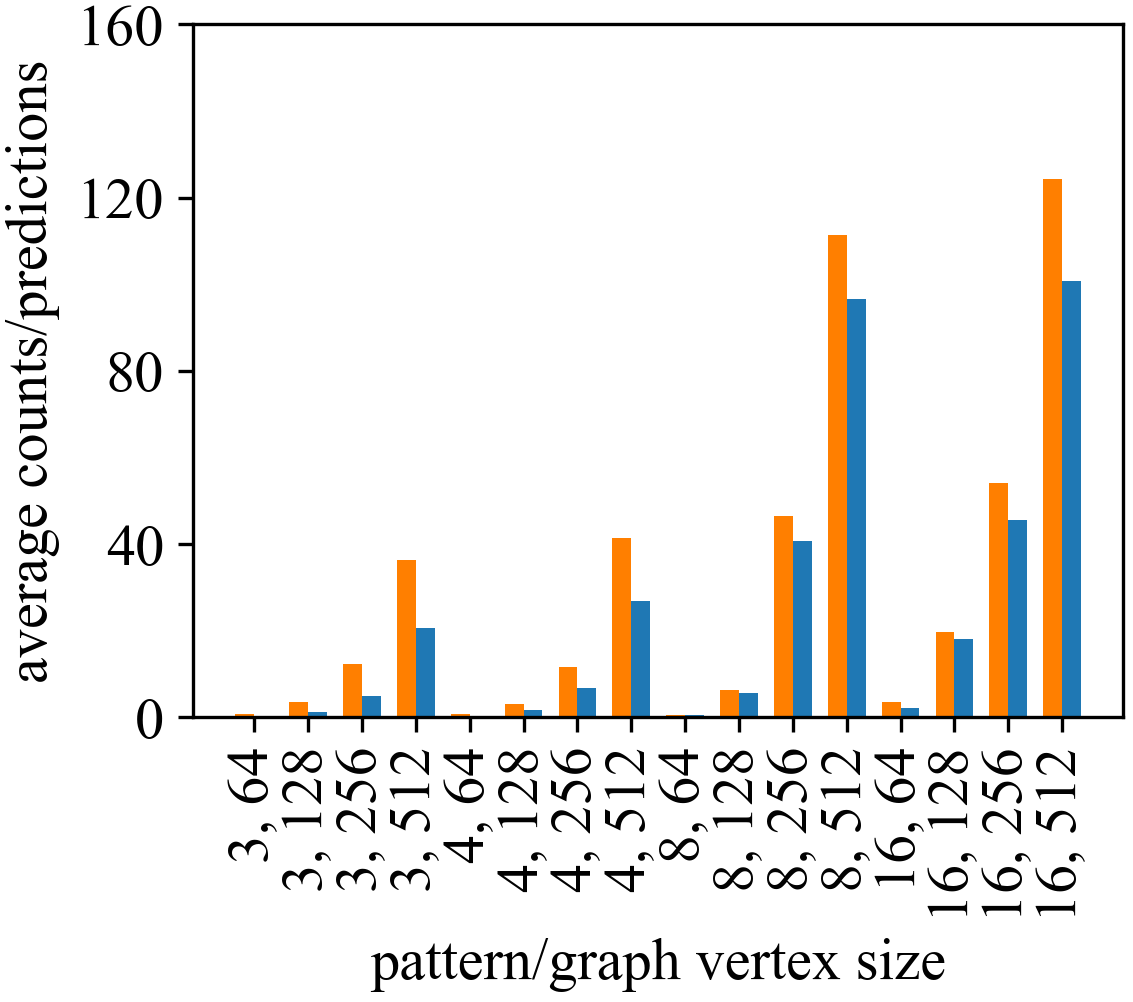}
        \caption{$O_\mathcal{V}$, CNN}
        \label{fig:CNN-DIAMNet-V-middle}
    \end{subfigure}
    \begin{subfigure}[b]{0.22\linewidth}
        \includegraphics[width=\linewidth,height=1.5in]{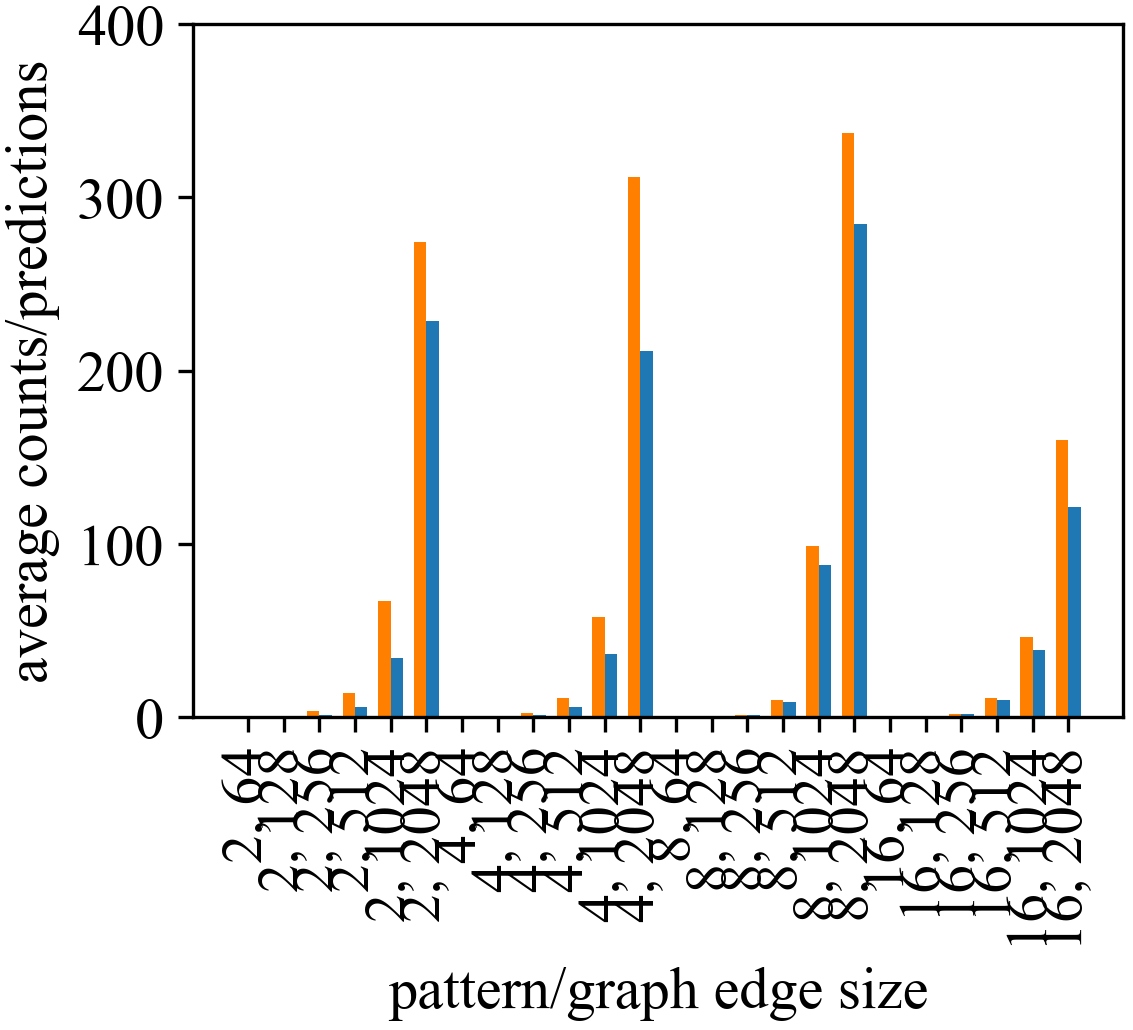}
        \caption{$O_\mathcal{E}$, CNN}
        \label{fig:CNN-DIAMNet-E-middle}
    \end{subfigure}
    \begin{subfigure}[b]{0.22\linewidth}
        \includegraphics[width=\linewidth,height=1.5in]{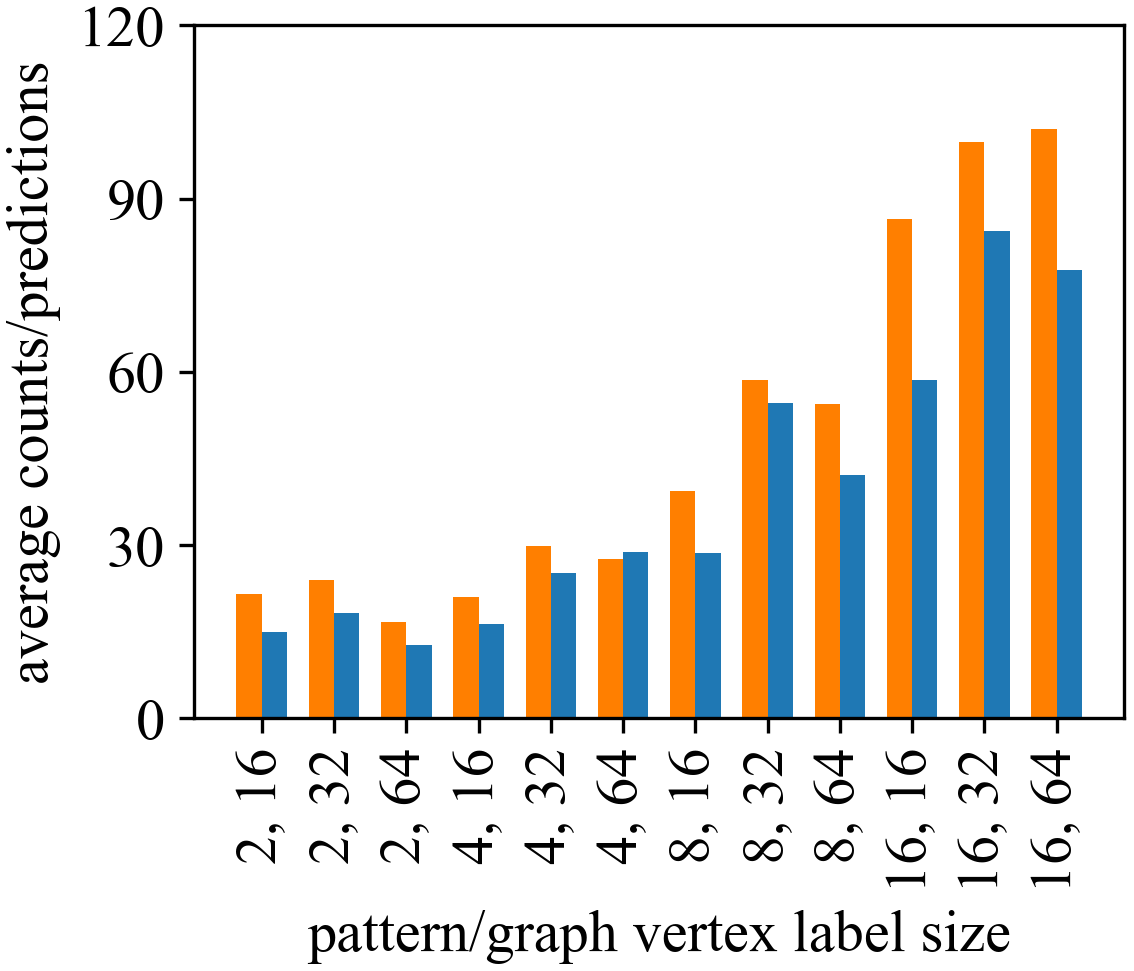}
        \caption{$O_\mathcal{X}$, CNN}
        \label{fig:CNN-DIAMNet-VL-middle}
    \end{subfigure}
    \begin{subfigure}[b]{0.22\linewidth}
        \includegraphics[width=\linewidth,height=1.5in]{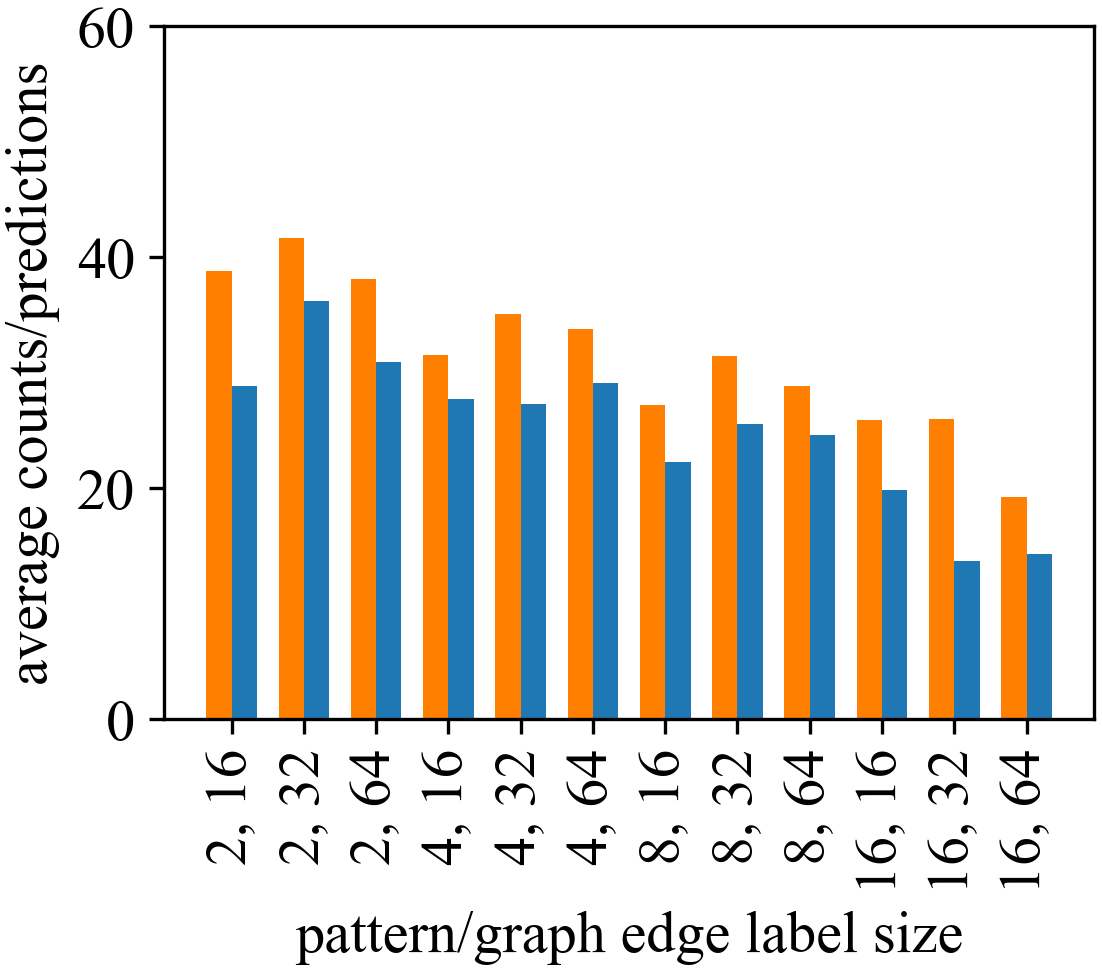}
        \caption{$O_\mathcal{Y}$, CNN}
        \label{fig:CNN-DIAMNet-EL-middle}
    \end{subfigure}	

	
    \begin{subfigure}[b]{0.22\linewidth}
        \includegraphics[width=\linewidth,height=1.5in]{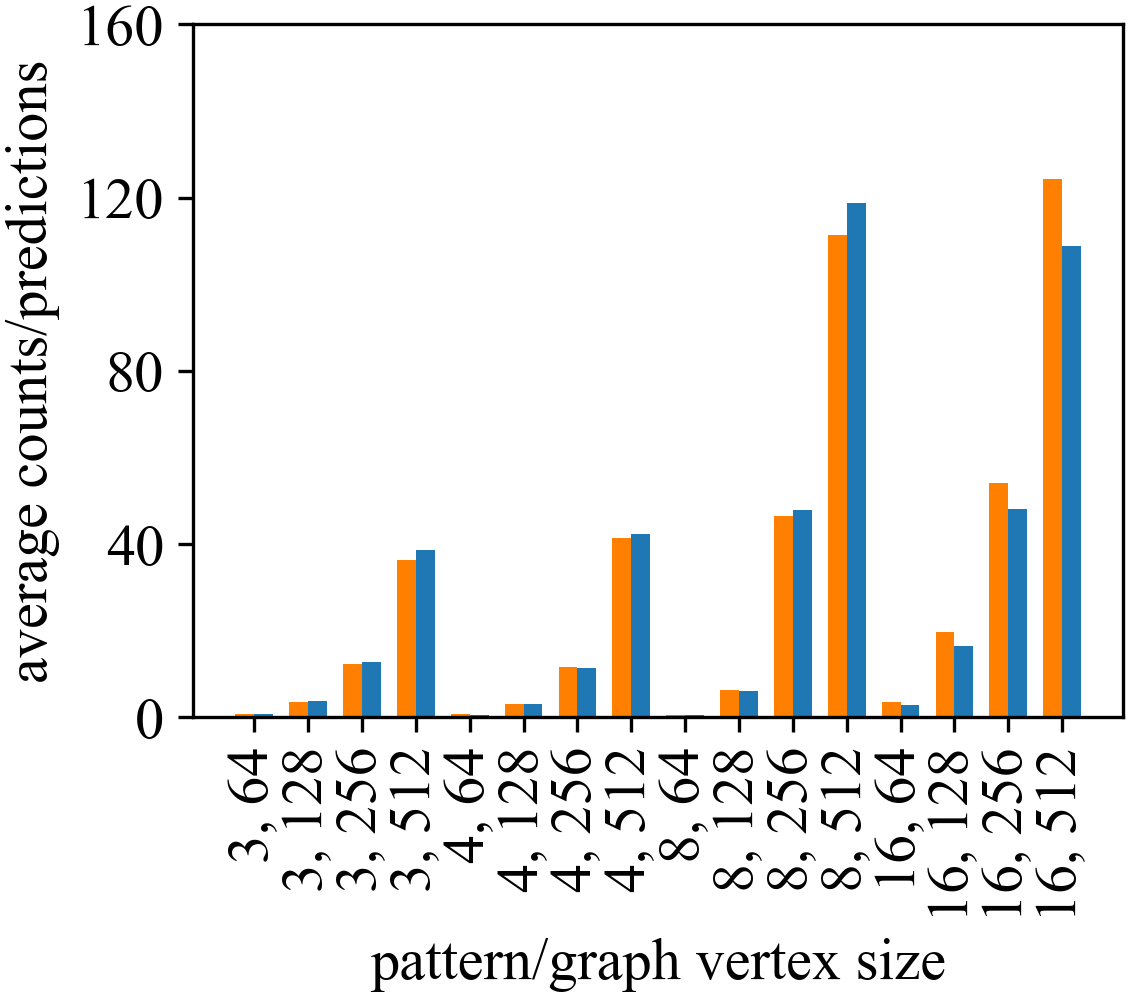}
        \caption{$O_\mathcal{V}$, RGIN}
        \label{fig:RGIN-DIAMNet-V-middle}
    \end{subfigure}
    \begin{subfigure}[b]{0.22\linewidth}
        \includegraphics[width=\linewidth,height=1.5in]{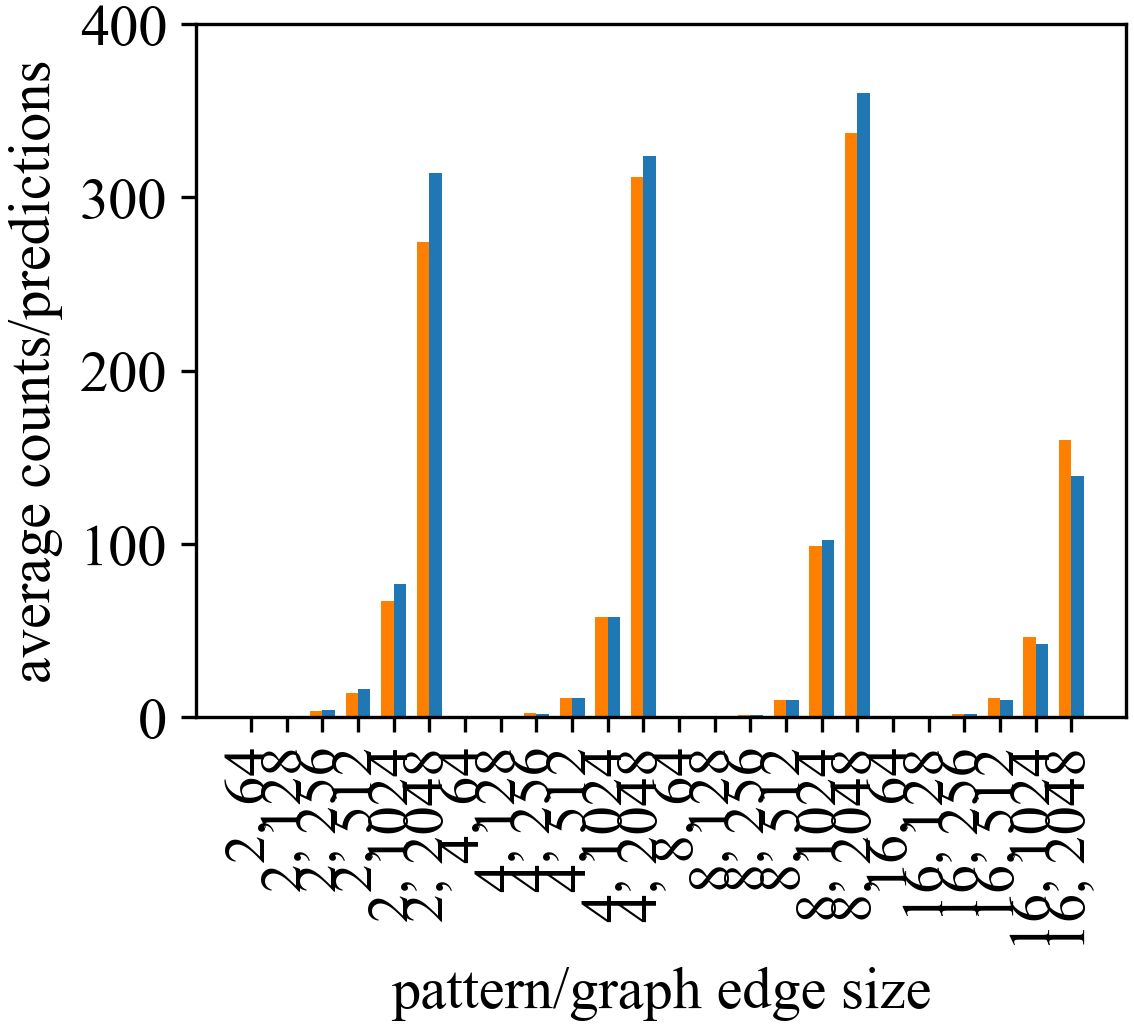}
        \caption{$O_\mathcal{E}$, RGIN}
        \label{fig:RGIN-DIAMNet-E-middle}
    \end{subfigure}
    \begin{subfigure}[b]{0.22\linewidth}
        \includegraphics[width=\linewidth,height=1.5in]{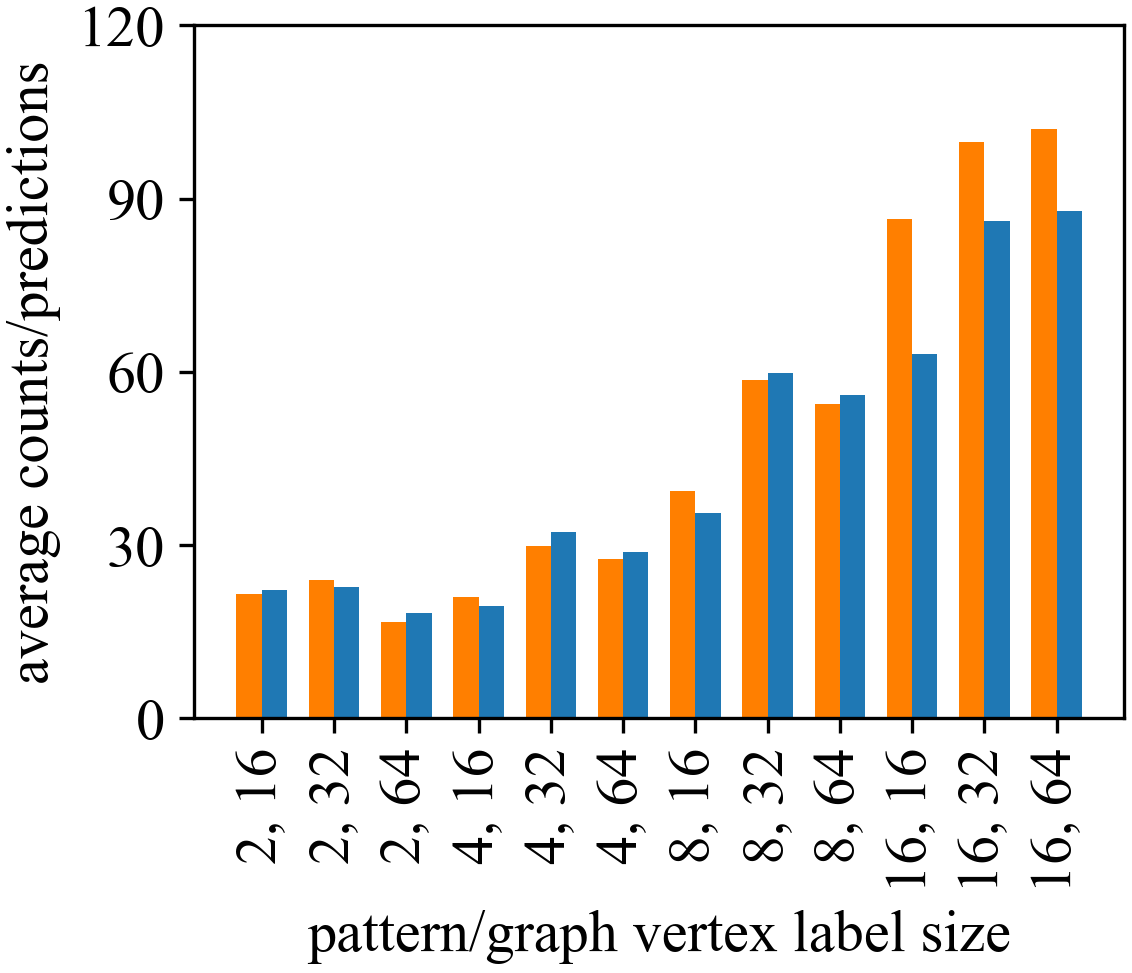}
        \caption{$O_\mathcal{X}$, RGIN}
        \label{fig:RGIN-DIAMNet-VL-middle}
    \end{subfigure}
    \begin{subfigure}[b]{0.22\linewidth}
        \includegraphics[width=\linewidth,height=1.5in]{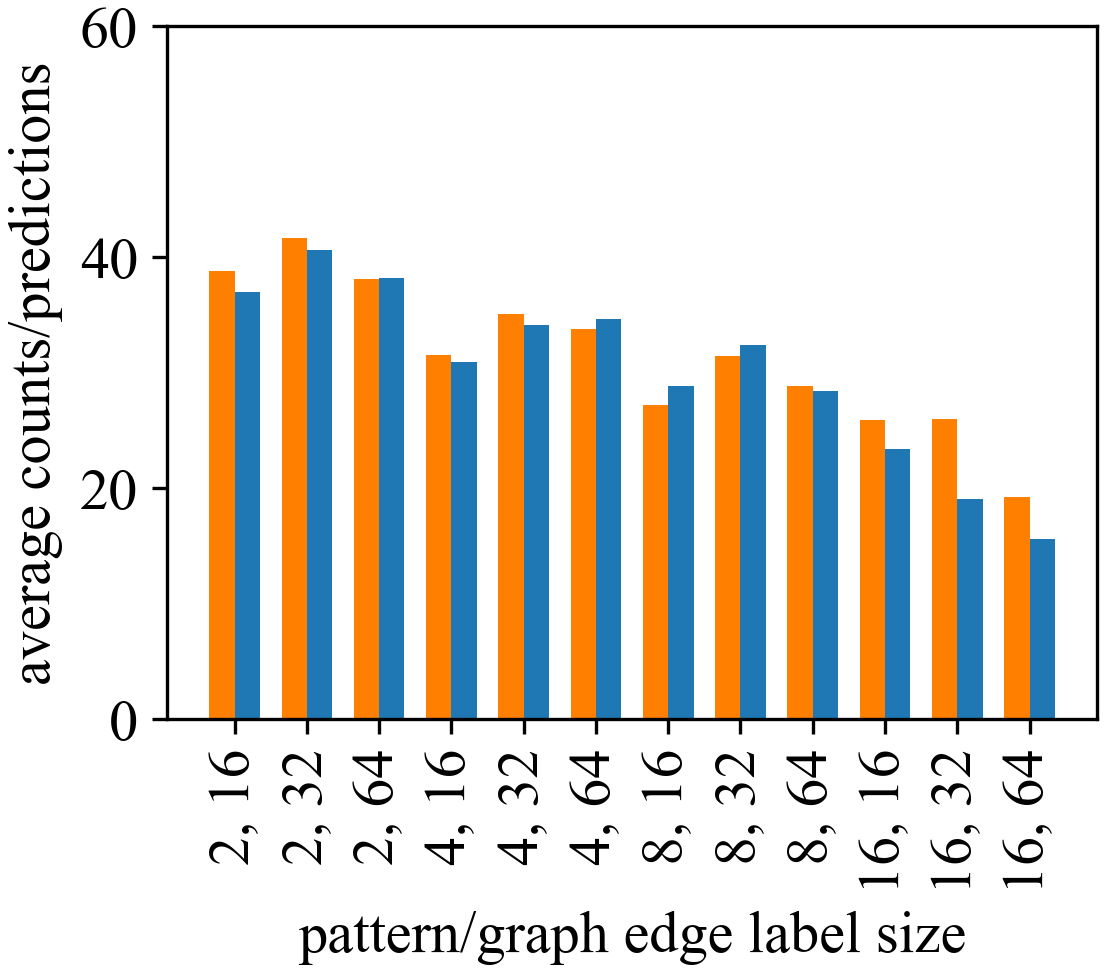}
        \caption{$O_\mathcal{Y}$, RGIN}
        \label{fig:RGCN-SUM_MeanDIAM_EL_middle}
    \end{subfigure}	
	\vspace{-0.1in}
    \caption{Model behaviors of CNN with DIAMNet and RGIN with DIAMNet on the \textit{large} dataset. Settings are similar to Figure~\ref{fig:model-behavior-small}.}\label{fig:model-behavior-middle}
    \vspace{-0.1in}
\end{figure*}

{\bf Effectiveness of the memory.} 
Table \ref{table:small_results} shows the effectiveness of DIAMNet as the prediction layer.
It outperforms the other pooling methods (SumPool, MeanPool, MaxPool) as well as the attention with memory mechanism (MemAttn) for all representation architectures. 
That means we can use a small extra GPU memory and corresponding computational cost to achieve better results.
The memory and computational cost of DIAMNet grows linearly with the size of inputs.
The efficiency can not be ignored when models are applied for real-life data with thousands or even millions of vertices and edges.
MemAttn can also reduce the quadratic computation cost to linear but its performance is not stable.

Prediction modules based on pooling are in general worse when the representation module is CNN or TXL. 
This observation indicates that context of the pattern representation should be counted and we need a better way to \revisexin{aggregate information from} context. 
DIAMNet can help extract the context of patterns and graphs even if the representation layer (such as CNN) does not perform very well. \revisexin{Finally, CNN with DIAMNet is the best among sequence models. The local information can be further used to infer in the global level with the help of memory.}

{\bf Performance on larger graphs.}
Table \ref{table:large_results} shows results of larger-scale graphs. 
We can find most of the results are consistent to the {\it small} dataset. RGIN is still the best representation method in our task and the dynamic memory also works effectively. 
In terms of the running time, all learning based models are much faster than the traditional VF2 algorithm for subgraph isomorphism counting.
\revisexin{Further more, the running time of neural models only increases linearly, making it possible to apply on large-scale real-life data.}

{\bf Model behaviors.}
As shown in Figure \ref{fig:model-behavior-small}, the performances of MaxPool and DIAMNet are compared by using the CNN models.

\revisexin{We can find MaxPool always predicts smaller subgraph isomorphisms, while DIAMNet can predict closer numbers even when graphs become large and complex because simple pooling methods will lose much information.
On the contrary, the DIAMNet uses small memory to learn which information should be reserved or dropped.}
In general, our DIAMNet has achieved the better results at most data scales. 
\revisexin{When bins are ordered by vertex label sizes or edge label sizes, we have the same observation that MaxPool fails when facing complex patterns with more distinct labels but DIAMNet can almost handle all complex situations.}

We can also conclude that different representation modules also perform differently from Figure \ref{fig:model-behavior-middle}. CNN performs worse when the graph size is large (shown in Figure \ref{fig:CNN-DIAMNet-V-middle} and \ref{fig:CNN-DIAMNet-E-middle}) and patterns become complicated (show in Figure \ref{fig:CNN-DIAMNet-VL-middle} and \ref{fig:CNN-DIAMNet-EL-middle}), which further indicates that CNN only extracts local information and suffers from inference of global information in larger graphs. 
On the contrary, RGIN with DIAMNet is not affected by edge sizes because it directly learns vertex representations rather than edge representations.
Moreover, vertex label sizes and edge label sizes have no apparent ill effects on RGIN, but CNN is seriously affected. 
Thus it can be seen that simply convolutional and pooling operations \revisexin{with little topology information} are not suitable for this task.

\subsection{Transfer Learning on \textit{MUTAG}}
\begin{figure}[t]
    \centering
    \includegraphics[width=0.8\linewidth]{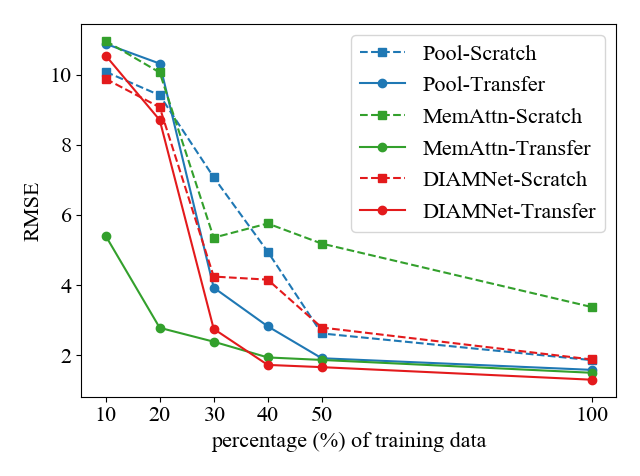}
    \vspace{-0.2in}
    \caption{Results of RGIN based models on the \textit{MUTAG} dataset. Dashed lines are results of models trained from scratch while solid lines are results by transfer learning.
    \revisexin{The \textbf{Zero} and the \textbf{Avg} result in 19.384 and 17.754 RMSE respectively when 100\% training data are used.}}\label{fig:mutag_results}
    \vspace{-0.2in}
\end{figure}

For the \textit{MUTAG} data, We feed different numbers of training pairs to RGIN with different interaction networks and try transfer learning by fine-tuning models trained on the \textit{small} dataset.
Results are shown in Figure~\ref{fig:mutag_results}.
The performance gap between models from scratch and models by transfer learning cannot be neglected when the training data are not enough.
With the help of transfer learning, the fine-tuned RGIN with any interaction modules can reduce RMSE smaller than 1.588 while the RMSE of the best model from scratch is still around 1.884.
\revisexin{When limited training data are given, all models perform not well except the MemAttn-Transfer. We can observe huge improvements by transfer learning when providing 30\%-40\% training data. DIAMNet-Transfer becomes the best afterwards and finally achieves 1.307 RMSE and 0.440 MAE.}
It is worth noticing that test graphs never appear in the training data on the \textit{MUTAG}, which indicates that our learning to count framework and algorithms are promising to handle large-scale real world graphs in the future.


\section{Related Work}

\noindent\textbf{Subgraph Isomophism Problems}. 
Given a pattern graph and a data graph, the subgraph isomorphism search aims to find all occurrences of the pattern in the data graph with bijection mapping functions. 
Subgraph isomorphism is an NP-complete problem among different types of graph matching problems (monomorphism, isomorphism, and subgraph isomorphism). 
Most subgraph isomorphism algorithms are based on backtracking. They first obtain a series of candidate vertices and update a mapping table, then recursively revoke their own subgraph searching functions to match one vertex or one edge at a time. 
Ullmann's algorithm~\citep{ullmann1976an}, VF2~\citep{vento2004a}, and GraphQL~\citep{HeS08} belong to this type of algorithms.
However, it is still hard to perform search when either the pattern or the data graph grows since the search space grows exponentially as well.
Some other algorithms are designed based on graph-index, such as gIndex~\citep{YanYH04}, which can be used as filters to prune out many unnecessary graphs. 
However, graph-index based algorithms have a problem that the time and space in indexing also increase exponentially with the growth of the graphs~\citep{SunWWSL12}.
TurboISO~\citep{Han2013turbo} and VF3~\citep{CarlettiFSV18} add some weak rules to find candidate subregions and then call the recursive match procedure on subregions. 
These weak rules can significantly reduce the searching space in most cases. 
\revisexin{\reviseyq{In addition,} there are also some approximation techniques~\cite{Tsourakakis08,EdenLRS17} for triangle counting but they are hard to generalized to large patterns. Random walks are also used to count subgraph isomorphisms~\cite{TeixeiraC0M18,WangLRTZG14} but the cost and errors are still issues.} 
\revisexin{Graph simulation~\cite{FanLMTWW10,MaCFHW14} is another way to find subgraph isomorphisms in cubic-time, but researchers require to rectify the failure of topology capturing and false or fake matches.}

\noindent\textbf{Graph Representation Learning}.
Graph (or network) representation learning can be directly learning an embedding vector of each graph node~\citep{PerozziAS14, TangQWZYM15, GroverL16}. This approach is not easy to generalize to unseen nodes. 
On the other hand, graph neural networks (GNNs)~\citep{abs-1806-01261} provide a solution to representation learning for nodes which can be generalized to new graphs and unseen nodes. 
Many graph neural networks have been proposed since 2005~\citep{Gori2005ANM,ScarselliYGHTM05} but rapidly developed in recent years. 
Most of them focus on generalizing the idea of convolutional neural networks for general graph data structures~\citep{NiepertAK16,kipf2017semi,HamiltonYL17,velivckovic2018graph} or relational graph structures with multiple types of relations~\citep{schlichtkrull2018modeling}.
More recently, \cite{XuHLJ19} propose a graph isomorphism network (GIN) and show its discriminative power. 
Others use the idea of recurrent neural networks (RNNs) which are originally proposed to deal with sequence data to work with graph data~\citep{LiTBZ15,YouYRHL18}.
Interestingly, with external memory, sequence models can work well on complicated tasks such as language modeling~\citep{sukhbaatar2015end,kumar2016ask} and shortest path finding on graphs~\citep{graves2016hybrid}.
There is another branch of research called graph kernels~\citep{VishwanathanSKB10,ShervashidzeSLMB11,YanardagV15,TogninalliNeurIPS2019,ChenNeurIPS2019} which also convert graph isomorphism to a similarity learning problem.
However, they usually work on small graphs and do not focus on subgraph isomorphism identification or counting problems.

\section{Conclusions}

In this paper, we study the challenging subgraph isomorphism counting problem.
With the help of deep graph representation learning, we are able to convert the NP-complete problem to a learning based problem.
Then we can use learned models to predict the subgraph isomorphism counts in polynomial time.
We build two \revisexin{synthetic} datasets to evaluate different representation learning models and global inference models.
Results show that learning based method is a promising direction for subgraph isomorphism detection and counting and \revisexin{our dynamic intermedium attention memory network} indeed help the global inference. 

\noindent {\bf Acknowledgements. }
This paper was supported by the Early Career Scheme (ECS, No. 26206717) and the Research Impact Fund (RIF, No. R6020-19) from the Research Grants Council (RGC) in Hong Kong,  and the Gift Fund from Huawei Noah's Ark Lab.

\bibliographystyle{ACM-Reference-Format}
\bibliography{kdd2020}

\clearpage

\clearpage
\appendix
\section{Lexicographic Order of Codes}
\label{appendix:lexicographic_order}
The lexicographic order is a linear order defined as follows: 

If $A=(a_0, a_1, \cdots, a_m)$ and $B=(b_0, b_1, \cdots, b_n)$ are the codes, then $A \leq B$ iff either of the following is true:
\begin{enumerate}
    \item $\exists t, 0 \leq t \leq min(m,n), \forall k<t, a_k = b_k, a_t \prec_e b_t$,
    \item $\forall 0 \leq k \leq m, a_k = b_k, \text{ and } n \geq m$.
\end{enumerate}

In our setting, $a_i=(u_{a_i}, v_{a_i}, \mathcal{X}(u_{a_i}), y_{a_i}, \mathcal{X}(v_{a_i})) \prec_e \\
b_j=(u_{b_j}, v_{b_j}, \mathcal{X}(u_{b_j}), y_{b_j}, \mathcal{X}(v_{b_j}))$ iff one of the following is true: 
\begin{enumerate}
    \item $u_{a_i} < u_{b_j}$,
    \item $u_{a_i} = u_{b_j}, v_{a_i} < v_{b_j}$,
    \item $u_{a_i} = u_{b_j}, v_{a_i} = v_{b_j}, y_{a_i} < y_{b_j}$.
\end{enumerate}

\section{Pattern Generator and Graph Generator}
\label{appendix:graph_generator}
\renewcommand{\algorithmicrequire}{\textbf{Input:}}
\renewcommand{\algorithmicensure}{\textbf{Output:}}

As proposed in Section~\ref{sec: dataset_generation}, two generators are required to generate datasets.
The algorithm about the pattern generator is shown in Algorithm~\ref{alg:pattern_generator}. The algorithm first uniformly generates a directed tree.
Then it adds the remaining edges with random labels.
Vertex labels and edge labels are also uniformly generated but each label is required to appear at least once.
Algorithm~\ref{alg:graph_generator} shows the process of graph generation. Two hyperparameters control the density of subisomorphisms: (1) $\alpha \in [0, 1]$ decides the probability of adding subisomorphisms rather than random edges; (2) $\beta \in \mathbb{N}^+$ is the parameter of Dirichlet distribution to sample sizes of components. 
After generating several directed trees and satisfying the vertex number requirement, the algorithm starts to add remaining edges.
It can add edges in one component and try to add subgraph isomorphisms, or it can randomly add edges between two components or in one component. The following merge subroutine aims to merge these components into a large graph. Shuffling is also required to make datasets hard to be hacked. The search of subisomorphisms in the whole graph is equivalent to the search in components respectively because edges between any two components do not satisfy the necessary conditions.

\begin{algorithm}[ht]
    \caption{Pattern Generator.}
    \label{alg:pattern_generator}
    {\footnotesize
    \begin{algorithmic}[1]
        \Require the number of vertices $N_v$, the number of edges $N_e$, the number of vertex labels $L_{v}$, the number of edge labels $L_{e}$.
        \State $\mathcal{P}$ := GenerateDirectedTree($N_v$)
        \State AssignNodesLabels($\mathcal{P}$, $L_{v}$)
        \State AddRandomEdges($\mathcal{P}$, $\mathcal{P}$, $null$, $N_{e}-N_v+1$) 
        \State AssignEdgesLabels($\mathcal{P}$, $L_{e}$)
        \Ensure the generated pattern $\mathcal{P}$
    \end{algorithmic}
    }
\end{algorithm}

\begin{algorithm}[ht!]
    \caption{Graph Generator.}
    \label{alg:graph_generator}
    {\footnotesize
    \begin{algorithmic}[1]
        \Require a pattern $\mathcal{P}$, the number of vertices $N_v$, the number of edges $N_e$, the number of vertex labels $L_v$, the number of edge labels $L_e$, hyperparameters $\alpha$ and $\beta$.
        \State $Ns$ := DirichletSampling($N_v$, $\beta$) 
        \State $\mathcal{G} s$ := $\{\}$
        \State $d_e$ := $N_e$
        \For {each $n$ in $Ns$}
            \State $g$ := GenerateDirectedTree($n$)
            \State AssignNodesLabels($g$, $n$)
            \State AssignEdgesLabels($g$, $n-1$)
            \State $\mathcal{G} s$ :=  $\mathcal{G} s + \{ g \}$
            \State $d_e$ := $d_e - n + 1$
        \EndFor
        \State $T_p$ := RewriteNECTree($\mathcal{P}$)
        \State $N_{e, \mathcal{P}}$ := EdgeCount($\mathcal{P}$)
        \While {$d_e > 0$}
            \State $g_1$, $g_2$ = RandomPick($\mathcal{G} s$, 2) \Comment{Pick two components randomly}
            \State $r$ = RandomNum(0, 1)
            \If {$d_e < N_{e, \mathcal{P}}$}
                \State $d_e$ := $d_e$ - AddRandomEdges($g_1$, $g_2$, $T_p$, $d_e$) \Comment{Add $d_e$ edges between $g_1$ and $g_2$}
            \Else
                \If {$r < \alpha$}
                    \State $d_e$ := $d_e$ - AddPatterns($g_1$, $\mathcal{P}$) \Comment{Add necessary edges in $g_1$ to add patterns}
                \Else 
                    \State $d_e$ := $d_e$ - AddRandomEdges($g_1$, $g_2$, $T_p$, $N_{e, \mathcal{P}}$)
                \EndIf
            \EndIf
        \EndWhile
        \State $\mathcal{G}$, $f$ := MergeComponents($\mathcal{G} s$) \Comment{$f$ is the graph id mapping}
        \State $\mathcal{G}$, $f'$ := ShuffleGraph($\mathcal{G}$) \Comment{$f'$ is the shuffled graph id mapping}
        \State $\mathcal{I} s$ := $\{ \}$
        \For {each $g$ in $\mathcal{G} s$}
            \State $Is$ := SearchSubIsomorphisms($g$, $\mathcal{P}$)
            \For {each $I$ in $Is$}
                \State $I$ := UpdateID($I$, $f$, $f'$)
                \State $\mathcal{I} s$ := $\mathcal{I} s + \{ I \}$
            \EndFor
        \EndFor
        \Ensure the generated graph $\mathcal{G}$, the subgraph isomorphisms $\mathcal{I} s$
    \end{algorithmic}
    }
\end{algorithm}

In Algorithm~\ref{alg:pattern_generator} and Algorithm~\ref{alg:graph_generator}, the function \textbf{AddRandomEdges} adds required edges from one component to the other without generating new subgraph isomorphisms. The two component can also be the same one, which means to add in one component. The NEC tree is utilized in TurboISO~\citep{Han2013turbo} to explore the candidate region for further matching. It takes $\mathcal{O}(|\mathcal{V}_p|^2)$ time but can significant reduce the searching space in the data graph. It records the equivalence classes and necessary conditions of the pattern. We make sure edges between two components dissatisfy necessary conditions in the NEC tree when adding random edges between them. This data structure and this idea help us to generate more data and search subisomorphisms compared with random generation and traditional subgraph isomorphism searching.

\section{Details of Datasets}
\subsection{Details of Synthetic Datasets}
\label{appendix:datasets}
We can generate as many examples as possible using  two graph generators.
However, we limit the numbers of training, dev, and test examples whether learning based models can generalize to new unseen examples.
Pattern and data graphs in different scales have been generated. 
Parameters for two generators and constraints are listed in Table~\ref{table:dataset_parameters}. 
Two constraints are added to make this task easier. The average degree constraint ($\frac{N_e}{N_v} \leq 4$) is used to keep graphs not so dense, and the subisomorphism counting constraints ($\{c \in \mathbb{N}| c \leq 1,024\}$ for \textit{small}, $\{c \in \mathbb{N}| c \leq 4,096\}$ for \textit{large}) ensure the difficulty of two datasets are different.

\begin{figure}[h]
    \centering
    \begin{subfigure}[b]{0.44\linewidth}
            \includegraphics[width=\linewidth]{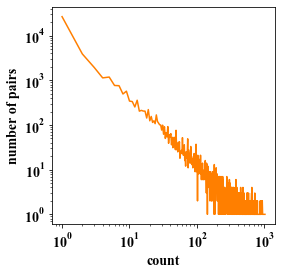}
            \caption{Test data distribution of the \textit{small} dataset.}
            \label{fig:small_distribution}
    \end{subfigure}
    \hspace{0.2in}
    \begin{subfigure}[b]{0.44\linewidth}
            \includegraphics[width=\linewidth]{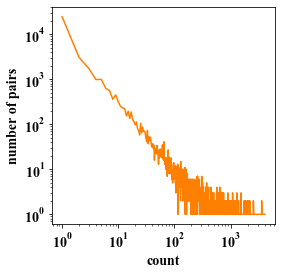}
            \caption{Test data distribution of the \textit{large} dataset.}
            \label{fig:large_distribution}
    \end{subfigure}
    \vspace{-0.1in}
    \caption{Subgraph isomorphism counting distributions of two datasets.}
    \label{fig:data_distribution}
\end{figure}

\revisexin{We generated data with an 8-core (16 threads) Intel E5-2620v4 CPU in parallel.}
The distributions of countings of two datasets are shown in Figure~\ref{fig:data_distribution}. From the figure we can see that both datasets follow the long tail distributions.

\begin{table}[h]
    \caption{Parameters and constraints for two datasets}
    \vspace{-0.1in}
    \label{table:dataset_parameters}
    \centering
    \footnotesize
    \begin{tabular}{l|c|c|c}
    \toprule
        & Hyperparameters & \textit{small} & \textit{large} \\
        \midrule
        \parbox[t]{2mm}{\multirow{4}{*}{\rotatebox[origin=c]{90}{Pattern}}}
        & $N_v$ & \{3, 4, 8\} & \{3, 4, 8, 16\} \\
        & $N_e$ & \{2, 4, 8\} & \{2, 4, 8, 16\} \\
        & $L_v$ & \{2, 4, 8\} & \{2, 4, 8, 16\} \\
        & $L_e$ & \{2, 4, 8\} & \{2, 4, 8, 16\} \\
        \hline
        \parbox[t]{2mm}{\multirow{6}{*}{\rotatebox[origin=c]{90}{Graph}}}
        & $N_v$ & \{8, 16, 32, 64\} & \{64, 128, 256, 512\} \\
        & $N_e$ & \{8, 16, $\cdots$, 256\} & \{64, 128, $\cdots$, 2,048\} \\
        & $L_v$ & \{4, 8, 16\} & \{16, 32, 64\} \\
        & $L_e$ & \{4, 8, 16\} & \{16, 32, 64\} \\
        & $\alpha$ & \{0.2, $\cdots$, 0.8\} & \{0.05, 0.1, 0.15\} \\
        & $\beta$ & \{512\} & \{512\}
        \\
        \bottomrule
    \end{tabular}
    \vspace{-0.1in}
\end{table}

\subsection{Details of \textit{MUTAG} Dataset}
\label{appendix:mutag}
\begin{figure}[t]
    \centering
    \includegraphics[width=0.8\linewidth]{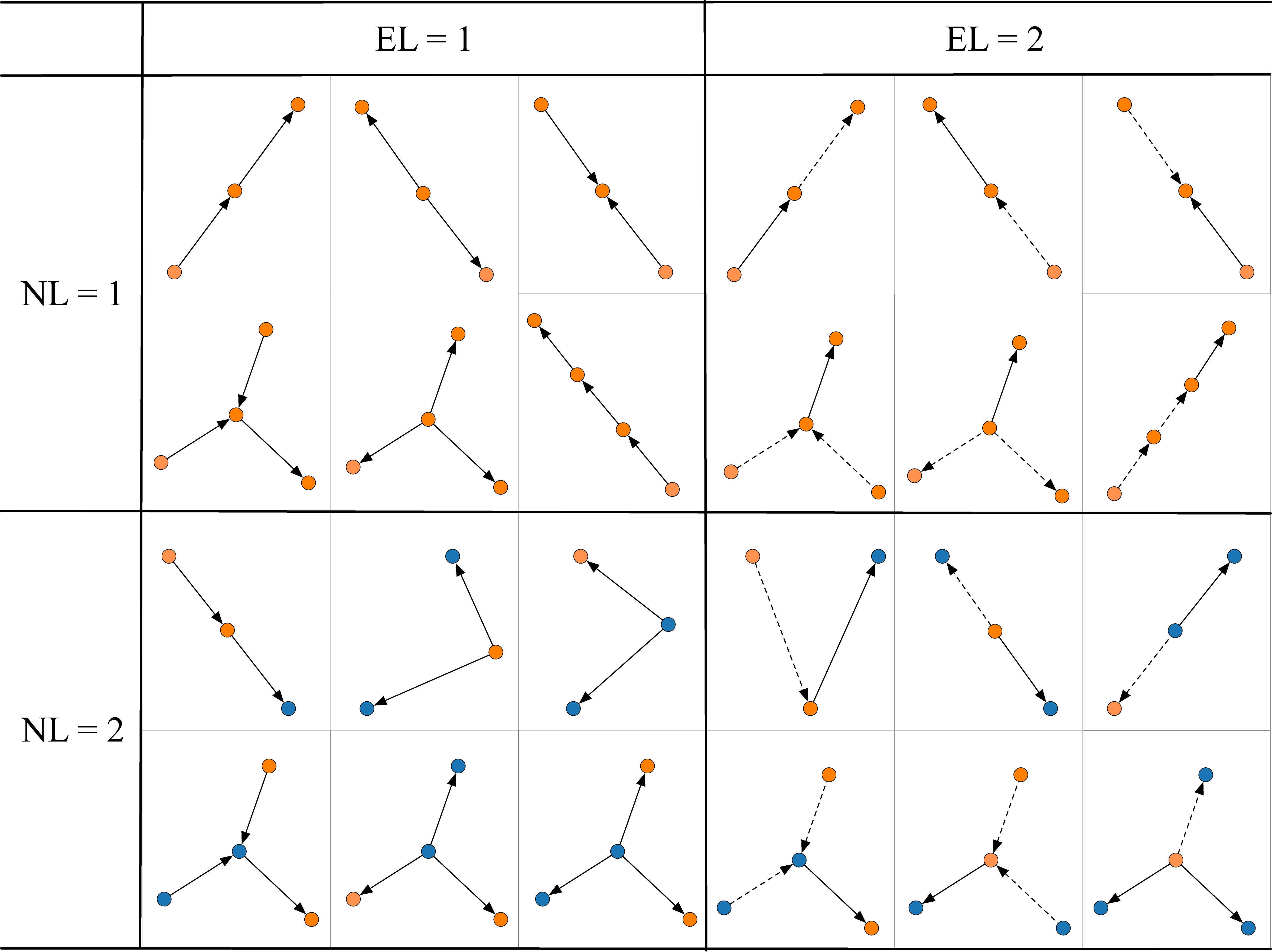}
    \vspace{-0.1in}
   \caption{24 different patterns in the \textit{MUTAG} dataset.}
    \label{fig:mutag_patterns}
\end{figure}

MUTAG dataset contains 188 mutagenic compound graphs.
We split 188 graphs as 62 training graphs, 63 validation graphs and 63 test graphs.
We use the pattern generator in Appendix~\ref{appendix:graph_generator} to generate 24 different patterns and finally construct 4,512 graph-pattern pairs (1,488 training pairs, 1,512 validation pairs and 1,512 test pairs).
The detailed pattern structures are shown in Figure~\ref{fig:mutag_patterns}.

\section{More Implementation Details}
In this section, we provide more implementation details other than the representation architectures and the DIAMNet modules.

\subsection{Filter Network}
\label{appendix:filternet}
Intuitively, not all vertices and edges in graphs can match certain subisomorphisms so we simply add a \textbf{FilterNet} to adjust graph encoding as follows:
{\small
\begin{align}
    \overline{\bm{p}} &= MaxPool(\bm{p}),  \label{eq:p_max} \\
    \widehat{\bm{G}} &= \bm{G} \bm{W}_G^{\top} \label{eq:g_layer},  \\
    \bm{f}_{i} &= \sigma(\bm{W}_F (\widehat{\bm{G}}_{i,:} \odot \overline{\bm{p}})) \label{eq:filter_gate},  \\
    \bm{G}_{i,:} &= \bm{f}_{i} \odot \bm{G}_{i,:},  \label{eq:filter}
\end{align}
}
where $\widehat{\bm{G}}$ is the graph representation in the pattern space, $\sigma$ is the sigmoid function, $f_{i}$ is the gate that decides to filter the $j^{th}$ vertex or the $j^{th}$ edge, $\bm{W}_G \in \mathbb{R}^{d_p \times d_g}$ and $\bm{W}_F \in \mathbb{R}^{1 \times d_p}$ are trainable. 
Thanks to the multi-hot encoding, We can simply use Eq.~(\ref{eq:p_max}) to accumulate label information of patterns.
After this filter layer, only relevant parts of the graphs will be passed to the next representation layer.

\subsection{MemAttn}
\label{appendix:memattn}
To address the quadratic-cost problem, we involve memory for the key and the value and do the dot-product between the query and the memory.
We initialize $\check{\bm{G}}^{(0)}$ as $\check{\bm{G}}$ and update $t$ times.
The details are listed as follows:
{\small
\begin{align}
    \bm{M}^{(t)} &= MemInit(\check{\bm{P}}), \\
    \bm{s}_j^{(t)} &= MultiHead(\check{\bm{g}}_j^{(t)}, \bm{M}^{(t)}, \bm{M}^{(t)}), \\
    \bm{z}_j^{(t)} &= \sigma(\bm{U}_P \check{\bm{g}}_j^{(t)} + \bm{V}_P \bm{s}_j^{(t)}), \\
    \overline{\bm{s}}_j^{(t)} &= \bm{z}_j^{(t)} \odot \check{\bm{g}}_j^{(t)} + (1 - \bm{z}_j^{(t)}) \odot \bm{s}_j^{(t)}, \\
    \overline{\bm{M}}^{(t)} &= MemInit(\overline{\bm{S}}^{(t)}), \\
    \widetilde{\bm{s}}_j^{(t)} &= MultiHead(\overline{\bm{s}}_j^{(t)}, \overline{\bm{M}}^{(t)}, \overline{\bm{M}}^{(t)}), \\
    \widetilde{\bm{z}}_j^{(t)} &= \sigma(\bm{U}_S \overline{\bm{s}}_j^{(t)} + \bm{V}_S \widetilde{\bm{s}}_j^{(t)}), \\
    \check{\bm{g}}_j^{(t + 1)} &= \widetilde{\bm{z}}_j^{(t)} \odot \overline{\bm{s}}_j^{(t)} + (1 - \widetilde{\bm{z}}_j^{(t)}) \odot \widetilde{\bm{s}}_j^{(t)}.
\end{align}
}
There are many ways to initialize memory, and we discuss  briefly in Appendix~\ref{appendix:meminit}.
In experiments, we simply choose to initialize by mean-pooling.


\subsection{Memory Initialization}
\label{appendix:meminit}
There are many ways to initialize the memory $\bm{M}=\{\bm{m}_1, \cdots, \bm{m}_M\}$ given an input $\bm{L}=\{\bm{l}_1, \cdots, \bm{l}_L\}$. 
In experiments, we simply initialize $\bm{m}_i$ by
{\small
\begin{align}
    s &= \lfloor \frac{L}{M} \rfloor,  \\
    k &= L - (M-1) \cdot s,  \\
    \bm{m}_i &= MeanPool(\{\bm{l}_{i \cdot s}, \cdots, \bm{l}_{i \cdot s+k-1}\}),  \label{eq:mean_attn_mem} 
\end{align}
}
where $s$ is the stride and $k$ is the kernel size.
Simply splitting the padded input with length $M \cdot \lceil \frac{L}{M} \rceil$ is also supposed to work well.
It is easy to replace the mean-pooling with max-pooling and sum-pooling.
Interestingly, we can also use self-attention for $\{\bm{l}_{i \cdot s}, \cdots, \bm{l}_{i \cdot s+k-1}\}$ to get a better memory block $\bm{m}_i$.
However, self-attention makes the initialization no faster than $\mathcal{O}(L^{1.5})$.
One further work is to design hashing and selection to help memory memorize better global and local structure information of $\bm{L}$, such as nodes with same vertex labels storing in the same memory block.

\end{document}